\pdfoutput=1

\documentclass[11pt]{article}

\usepackage[preprint]{acl}

\usepackage{times}
\usepackage{latexsym}
\usepackage[T1]{fontenc}
\usepackage[utf8]{inputenc}
\usepackage{microtype}
\usepackage{inconsolata}
\usepackage{graphicx}

\usepackage{amsmath}
\usepackage{amssymb}
\usepackage{booktabs}
\usepackage{array}
\usepackage{multirow}
\usepackage{algorithm}
\usepackage{algpseudocode}
\usepackage{xspace}
\usepackage{tikz}
\usetikzlibrary{positioning,arrows.meta}

\definecolor{decred}{RGB}{190,40,40}
\definecolor{firstorange}{RGB}{235,140,30}
\colorlet{agentA}{blue}
\colorlet{agentB}{orange}
\colorlet{agentC}{green!60!black}

\definecolor{colfaaccent}{HTML}{D97757}
\definecolor{colfatext}{HTML}{C4613F}
\colorlet{colfarow}{colfaaccent!12}
\newcommand{\shadeht}{\dimexpr\ht\strutbox+\dp\strutbox+\aboverulesep+\belowrulesep\relax}
\newcommand{\shaderow}[1]{\noalign{\begingroup\color{#1}\hrule height \shadeht\endgroup\vskip-\shadeht}}
\newcommand{\shadestrut}{\rule{0pt}{\dimexpr\ht\strutbox+\belowrulesep\relax}\rule[-\dimexpr\dp\strutbox+\aboverulesep\relax]{0pt}{0pt}}
\newlength{\numheadw}
\newcommand{\numhead}[3][c]{\settowidth{\numheadw}{#2}\makebox[\numheadw][#1]{#3}}
\newcommand{\shadestrutht}{\rule{0pt}{\dimexpr\ht\strutbox+\belowrulesep\relax}}
\newcommand{\shadestrutend}{\rule[-\dimexpr\dp\strutbox+\aboverulesep\relax]{0pt}{0pt}}
\newcommand{\shadelines}[2]{\noalign{\begingroup\color{#1}\hrule height \dimexpr#2\ht\strutbox+#2\dp\strutbox+\aboverulesep+\belowrulesep\relax\endgroup\vskip-\dimexpr#2\ht\strutbox+#2\dp\strutbox+\aboverulesep+\belowrulesep\relax}}
\newsavebox{\numcellbox}
\newcolumntype{N}[1]{>{\begin{lrbox}{\numcellbox}}c<{\end{lrbox}\settowidth{\numheadw}{#1}\makebox[\numheadw][r]{\usebox{\numcellbox}}}}
\newcommand{\hc}[1]{\multicolumn{1}{c}{#1}}

\hypersetup{pdftitle={CoLFA: Collaboration-Guided Lightweight Failure Attribution for LLM-Based Multi-Agent Systems},
  pdfauthor={Yufei Luo, Yuchen Ling, Chunrong Fang}}
\usepackage[skins,breakable]{tcolorbox}

\newcounter{promptlisting}
\renewcommand{\thepromptlisting}{\arabic{promptlisting}}

\newenvironment{prompttemplate}[2][]{%
  \refstepcounter{promptlisting}%
  \begin{tcolorbox}[
    enhanced,
    colback=gray!3,
    colframe=black!55,
    coltitle=white,
    colbacktitle=black!80,
    title={Prompt~\thepromptlisting: #2},
    fonttitle=\bfseries,
    boxrule=0.5pt,
    arc=1mm,
    left=7pt,
    right=7pt,
    top=7pt,
    bottom=7pt,
    boxsep=1pt,
    toptitle=3pt,
    bottomtitle=3pt,
    lefttitle=7pt,
    righttitle=7pt,
    before upper={
      \small
      \linespread{1.1}\selectfont
      \setlength{\parindent}{1pt}
      \setlength{\parskip}{1pt}
      \raggedright
    },
    #1
  ]%
}{%
  \end{tcolorbox}%
}

\newcommand{\promptsection}[1]{%
  \par\vspace{4pt}%
  \noindent\textbf{#1}\par\vspace{2pt}%
}

\graphicspath{{figures/}}

\newcommand{\method}{CoLFA\xspace}

\title{CoLFA: Collaboration-Guided Lightweight Failure Attribution for LLM-Based Multi-Agent Systems}

\author{Yufei Luo \quad Yuchen Ling \quad Chunrong Fang\thanks{Corresponding author.} \\
  Nanjing University \\
  \texttt{luoyufei0510@gmail.com}, \texttt{yuchenling@smail.nju.edu.cn}, \texttt{fangchunrong@nju.edu.cn}}

\begin{document}
\maketitle

\begin{abstract}

Failure attribution in multi-agent systems requires tracing error formation to identify the decisive error step and the responsible agent. However, agents' division of work and information exchange lead to Interleaved Steps and Cluttered Content, requiring models to link executions and check details in lengthy records. This makes it difficult to balance localization accuracy and token cost. We find that multi-agent collaboration also leaves records of how executions connect and content passes between them as results are formed, providing evidence for tracing error formation. Based on this observation, we propose CoLFA (Collaboration-Guided Lightweight Failure Attribution), which uses collaboration cues to organize the attribution input according to how results are formed, determining which steps are examined and how their content is presented. Through Handoff-Guided Step Expansion and Cross-Referenced Content Organization, CoLFA helps the model focus on examining error formation and the role of each decision while reducing the processing of irrelevant and repeated records. Across five settings spanning Who\&When and LongRCA and three backbone models, CoLFA achieves the highest exact step-localization accuracy with the lowest token consumption. Compared with individual baselines in each setting, CoLFA improves accuracy by up to 37.50\% and reduces token consumption by up to 97.6\%.\footnote{\raggedright Code will be released at \url{https://github.com/luoyufeihaha/CoLFA}.}

\end{abstract}

\section{Introduction}

\label{sec:intro}

\begin{figure}[t]
  \centering
  \includegraphics[width=\columnwidth]{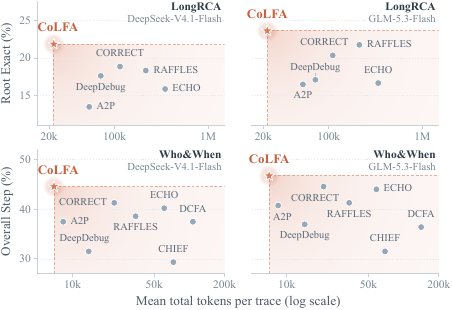}
  \caption{Exact step-localization accuracy and token consumption on Who\&When and LongRCA.}
  \label{fig:cost-accuracy}
\end{figure}

As LLM-based multi-agent systems (MAS) are increasingly applied to complex, long-horizon tasks, obtaining useful feedback from failed executions has become important for system debugging and continuous improvement \citep{longrca,learnfail}. Failure attribution aims to identify the decisive error step, defined as the earliest erroneous step that plays a decisive role in the final failure, and the responsible agent \citep{zhang2025agentcausestaskfailures}.

The final result of a multi-agent system is formed progressively through the division of work among agents, handoffs, and subsequent processing. An error may be introduced upstream and become apparent only after further transfer and processing, or it may arise from downstream interpretation or processing. Earlier errors may also be corrected during subsequent execution. Thus, erroneous content appearing at a step does not necessarily mean that the decisive error occurred there. Failure attribution requires tracing how the final erroneous result was formed, linking related executions to examine where errors were introduced, how they changed, and what role they played in the final failure \citep{chief,longrca}.

In multi-agent systems, local executions are relatively independent, while interactions across agents connect them within the overall task \citep{autogen,metagpt}. This form of collaboration poses two related challenges for failure attribution when tracing how results are formed. The first challenge, \textbf{Interleaved Steps}, arises because steps involving task requirements, result production, and subsequent processing may belong to different agents, span multiple execution stages, and be interleaved with other work. The chronological order of the original execution trace does not readily reveal these connections, making it difficult for the model to link earlier and later executions. The second challenge, \textbf{Cluttered Content}, arises because the evidence needed for attribution is mixed with extensive reasoning, actions, and environment feedback from different agents. Relevant content may also be repeated, paraphrased, or modified as it passes across steps. This makes key details harder to identify and compare, while long and repetitive inputs increase token cost.

To address these challenges, prior work uses guided reasoning, structured context, candidate search, and verification to strengthen the analysis of execution traces. One group of methods directly analyzes the original execution trace, with relatively low token cost but limited localization accuracy \citep{a2p,correct}. Another group uses structured processing or multiple rounds of analysis to examine executions more thoroughly \citep{echo,raffles,chief}. In the comparison in Figure~\ref{fig:cost-accuracy}, however, these methods incur much higher token costs with relatively limited localization gains. The presence of evidence in the input does not mean that it is easy to connect and check. Both groups still rely on model analysis to varying degrees. Can we overcome the trade-off between localization accuracy and analysis cost by locating decisive error steps more accurately while reducing token cost?

We find that the same collaborative process that gives rise to Interleaved Steps and Cluttered Content also leaves evidence for linking executions and comparing earlier and later processing. \textbf{At the step level,} determining whether an error occurred in a particular execution requires considering its task requirements and inputs, which may be distributed across other agents' records. Handoff records capture the transfer of requirements and intermediate results between agents, providing clues for finding the corresponding source records. Content correspondence across steps helps connect the production and subsequent processing of the same information. These cues help link executions that need to be examined together, even before the error location is known. How much historical context is needed depends on what each record helps check.

\textbf{At the content level,} the same content may be handed off, referenced, or used at different steps, with its recurrence in execution records providing content correspondence across steps. This content provides a shared point of comparison for checking the basis on which it was produced and how it was subsequently processed. Task requirements and available information may differ across executions, so each instance of processing needs to be assessed in its own context to determine whether it is erroneous. We refer to handoff records and cross-step content correspondence in execution traces as \textbf{collaboration cues}. Evidence for attribution should therefore be organized according to how the result is formed, so that related executions can be examined together and each decision can be assessed in its own context. Setting the scope and level of detail of records according to what needs to be checked can make evidence easier to check while reducing unnecessary expansion and repeated presentation.

Based on these observations, we propose \textbf{CoLFA (Collaboration-Guided Lightweight Failure Attribution)}, which uses collaboration cues to organize the attribution input according to how the result is formed. Its two complementary components determine which steps are examined and how their content is presented. \textbf{Handoff-Guided Step Expansion} starts from the final result, links executions through handoffs and content correspondence, and sets each agent's expansion scope according to what needs to be checked. \textbf{Cross-Referenced Content Organization} uses references to establish a shared point of comparison and retains a separate record for each execution, enabling the model to compare earlier and later processing and assess each decision in its own context. This organization helps the model focus on examining error formation and the role of each decision in the final failure, while reducing the processing of irrelevant history and repeated text. It thereby improves decisive error localization accuracy and lowers token cost.

In evaluations on all failed execution traces from \textbf{Who\&When} and \textbf{LongRCA} using two backbone models, CoLFA achieves the highest exact step-localization accuracy with the lowest token consumption in all four settings. Compared with the highest-accuracy baseline in each setting, CoLFA improves accuracy by 1.84\% to 3.26\% while reducing token consumption by 66\% to 90\%.

\section{Related Work}

\label{sec:related}

\subsection{Multi-Agent Failure Attribution}

Failure attribution in multi-agent systems aims to identify the decisive error step and the responsible agent, providing evidence for understanding and repairing system failures \citep{zhang2025agentcausestaskfailures,longrca}. Existing methods mainly follow two strategies. In one strategy, the model analyzes the complete execution trace in a single pass, using guided reasoning \citep{a2p} or historical error patterns \citep{correct} to support its judgment. These methods have relatively low token cost, but localization accuracy remains limited on long, cluttered records. The other strategy uses candidate search, investigation, and verification to examine suspicious steps more thoroughly \citep{raffles,agentdebugx,falat,dcfa,chief}. These methods require additional model analysis during search and verification, resulting in higher token costs with relatively limited gains in localization accuracy, as shown in Figure~\ref{fig:cost-accuracy}. To balance localization accuracy and token cost, CoLFA organizes the attribution input to help the model focus on examining how errors are formed and reduce the processing of irrelevant and repeated records.

\subsection{Trace Organization for Attribution}

Trace organization determines which execution records the attribution model receives and how their content is presented, affecting its ability to link earlier and later executions and check evidence \citep{aig,searchauditor,codetracer}. Prior work mainly improves attribution analysis by controlling input size or organizing relations between executions. One approach reduces long inputs by compressing or filtering execution records \citep{echo,duotrace,trajdebug}, or retrieves original text and reads it on demand to provide the context needed for examination \citep{trajaudit,safari,longrca}. They control how much the model reads, but compression and filtering may omit details needed for examination, and supplying context on demand requires further retrieval and reading. The other approach builds dependency structures from input-output correspondence \citep{cdcmas} or model inference \citep{chief,aig} to support tracing errors along relations between executions \citep{agenttrace}. For methods that rely on the model to build these structures, performance depends on the accuracy of relation identification. Building and checking these structures also incur additional analysis cost. To control input size while supporting comparisons between earlier and later processing, CoLFA starts from the final result, links related executions through handoffs and content correspondence, uses references to establish a shared point of comparison, and retains a separate record for each execution.

\section{Method}

\label{sec:method}

\begin{figure*}[t]
  \centering
  \includegraphics[width=\textwidth]{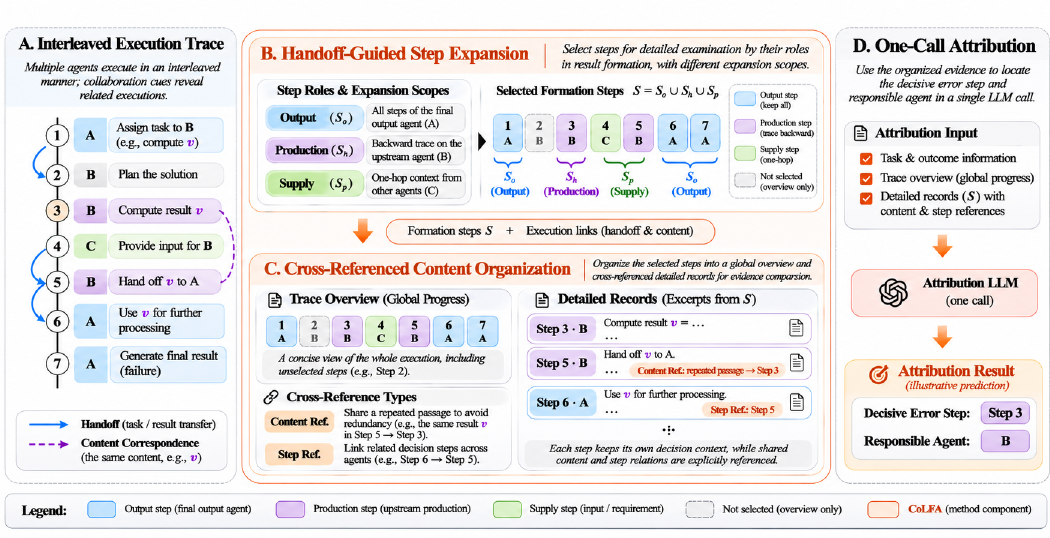}
  \caption{Overview of \method, showing how collaboration cues guide step expansion and content organization.}
  \label{fig:overview}
\end{figure*}

\subsection{Task Definition and Method Overview}

\label{sec:method-overview}

Given a task description \(q\) and a failed execution trace \(\tau=(s_1,\ldots,s_T)\), failure attribution aims to identify the decisive error step and the responsible agent. The trace contains the agents' decision steps \(s_i\) in chronological order, with \(a_i\) denoting the author of step \(s_i\). Step indices retain the numbering in the original execution record. Tool outputs are included in the step that issued the corresponding call. Each step record may contain reasoning, an action, and tool outputs or environment feedback.

As shown in Figure~\ref{fig:overview}, CoLFA organizes the attribution input based on how the final result is formed. Using collaboration cues, it determines which steps are examined and how their content is presented to address Interleaved Steps and Cluttered Content. \textbf{Handoff-Guided Step Expansion (Section~\ref{sec:method-expansion})} brings related executions together, and \textbf{Cross-Referenced Content Organization (Section~\ref{sec:method-organization})} enables the model to compare information processing across these executions. The model uses the organized records and overall execution progress to identify the decisive error step among the selected steps and the responsible agent.

\subsection{Handoff-Guided Step Expansion}

\label{sec:method-expansion}

Given an execution trace \(\tau\), let \(V\) be the set of original indices of all decision steps. Starting from the final result, CoLFA uses collaboration cues at the step level to select steps for examining how the result was formed. We call these formation steps and denote their index set by \(S\subseteq V\). They bring together executions by different agents at different points in time, providing evidence for the model to locate the decisive error step.

Handoff records capture the transfer of task requirements and intermediate results, while content correspondence suggests that different executions may involve the same information. These cues help trace related executions backward from the final result. After establishing these links, we determine how much history to include, since tracing how content was produced and checking requirements and inputs require different contexts. To address Interleaved Steps, we use collaboration cues to link related executions and set each agent's expansion scope according to what needs to be checked.

\looseness=1
For each step \(i\), we identify earlier steps linked to it through handoffs or content correspondence. We denote the set of their indices by \(P(i)\), which represents the execution links of step \(i\). Handoff cues include assignments, mentions of an agent, and changes in the speaking agent. Content correspondence is identified from recurring numerical values, identifiers, or specific text across steps, excluding content already given in the task and common content. Detailed identification rules are provided in Appendix~\ref{sec:appendix-links}. For collaboration configurations in which agents delegate and hand off work, we organize \(S\) into output steps, production steps, and supply steps according to what needs to be checked. These respectively support examining the output agent's processing in sequence, tracing how handed-off content was produced upstream, and checking the processing against requirements and \mbox{inputs supplied by other agents}.

\paragraph{Output steps.}
The output agent may use information, take actions, and make revisions over multiple steps to form the final result, so its processing needs to be examined in sequence. We define the output agent \(a_o\) as the author of the final decision step. We retain all of its decision steps throughout the trace and denote their index set by \(S_o\).

\paragraph{Production steps.}
To check whether an error was introduced upstream, we select the author of the last decision step not taken by the output agent as the upstream agent \(a_h\). We then trace how the handed-off content was produced. Let \(P_c(i)\) be the subset of \(P(i)\) obtained through content correspondence. We initialize and recursively expand the index set of production steps as
\begingroup
\predisplaypenalty=0
\begin{align}
\label{eq:production-steps}
S_h^{(0)}
&=
\{
j\in\bigcup_{i\in S_o}P(i)
\mid  a_j=a_h
\},\\
S_h^{(r+1)}
&=
S_h^{(r)}
\cup
\{
j\in\bigcup_{i\in S_h^{(r)}}P_c(i)
\mid  a_j=a_h
\}.\notag
\end{align}
\endgroup
The initial set contains the upstream steps linked to any output step through handoffs or content correspondence. Subsequent iterations trace backward within the same agent using only content correspondence. \(r\) denotes the iteration number, and iteration continues until the set stabilizes, yielding \(S_h\).

\paragraph{Supply steps.}
When examining the output and production steps, we need the requirements and inputs supplied by the remaining agents. To provide context while controlling how much history is included, we define the index set of supply steps as
\begin{equation}
\label{eq:supply-steps}
S_p=
\{
j\in\bigcup_{i\in S_o\cup S_h}P(i)
\mid
a_j\notin\{a_o,a_h\}
\}.
\end{equation}
\(S_p\) contains only steps directly linked to output or production steps and authored by the remaining agents. Expansion stops at these supply steps.

Together, the three parts define \(S = S_o \cup S_h \cup S_p\), the index set of formation steps. We adjust the expansion scopes according to the known collaboration configuration, as detailed in Appendix~\ref{sec:appendix-scope}.

\subsection{Cross-Referenced Content Organization}

\label{sec:method-organization}

Given the task description \(q\), execution trace \(\tau\), and formation steps \(S\), CoLFA uses collaboration cues at the content level to organize detailed records of these steps. It combines the task description, detailed records, and trace overview into the attribution input. This input enables the model to compare how information is processed across executions and provides overall execution progress to help locate the decisive error step.

Content can be reused across steps, but each decision involving it has its own execution context. To address Cluttered Content, we retain a separate record for each execution and use shared content and execution links to support comparisons between earlier and later processing. We also incorporate overall execution progress to help the model assess the role of errors in the final failure.

\paragraph{Detailed records.}
We retain a separate record for each formation step, specifying its original index and author and presenting its reasoning, action, and tool outputs or environment feedback. Content references and step references support comparisons between these records. Content references point to earlier occurrences of repeated text, allowing the model to compare earlier and later processing of the same content. Step references identify relevant earlier steps so that the model can check the current execution against the requirements, inputs, and processing recorded there. Actions are not truncated for length. We truncate reasoning, tool outputs, or environment feedback when they are overly long and mark the omissions. Detailed settings are provided in Appendix~\ref{sec:appendix-input}.

\paragraph{Content references.}
To compare earlier and later processing of the same content, content references allow repeated text to be shared across records, providing a shared point of comparison and reducing repeated presentation. We identify contiguous repeated regions from identical word sequences shared with the task description or previously presented detailed records. We replace these regions with references to their primary matching sources, following the rules given in Appendix~\ref{sec:appendix-input}. The model uses the records to compare and assess paraphrases and modifications.

\paragraph{Step references.}
To examine the output process and upstream content production, the model needs to revisit relevant requirements, inputs, and earlier processing. Using the execution links identified during step expansion, we define the referenced step indices for each \(i\in S_o\cup S_h\) as
\begin{equation}
\label{eq:step-references}
R(i)=\{j\in P(i)\cap S\mid a_j\ne a_i\}.
\end{equation}
The index and author of each step in \(R(i)\) are listed in the detailed record of step \(i\), providing explicit pointers for checking relevant requirements, inputs, and earlier processing.

\paragraph{Trace overview.}
Assessing an error's role in the final failure requires examining subsequent processing, which may occur outside the formation steps. The trace overview extracts key points of actions and tool outputs or environment feedback across the complete execution and presents brief chronological records labeled with their original step indices and authors. It provides overall execution progress as context for examining the processing in the detailed records. The construction procedure is given in Appendix~\ref{sec:appendix-input}.

The model receives the task description, trace overview, and detailed records as the attribution input. It judges whether an action was erroneous based on the agent's task requirements and the information available at that step, and uses subsequent processing to assess the error's role in the final failure. In a single call, it identifies the decisive error step among the formation steps and returns its author as the predicted responsible agent.

\section{Experiments}

\label{sec:experiments}

\subsection{Experimental Setup}

\label{sec:exp-setup}

\paragraph{Benchmarks.}
We evaluate on all failed execution traces from Who\&When~\citep{zhang2025agentcausestaskfailures} and LongRCA~\citep{longrca}, which cover different trace lengths and collaboration configurations. Who\&When comprises Algorithm-Generated (Alg.) and Hand-Crafted (Hand.) splits, containing 126 and 58 traces with average lengths of 8.7 and 51.6 steps, respectively. LongRCA contains 1,140 traces with an average length of 156.0 steps and covers multiple collaboration configurations. We provide the attribution model with task reference answers on Who\&When but not on LongRCA. Step counts are based on the original execution records. Benchmark details and input protocols are provided in Appendix~\ref{sec:appendix-benchmarks}.

\paragraph{Baselines.}
We compare against ten baselines covering two attribution strategies. The first group directly analyzes complete execution traces and includes All-at-Once~\citep{zhang2025agentcausestaskfailures}, A2P~\citep{a2p}, and CORRECT~\citep{correct}. The second performs attribution through step-by-step examination, candidate search, or multiple rounds of analysis and includes Step-by-Step and Binary Search~\citep{zhang2025agentcausestaskfailures}, ECHO~\citep{echo}, RAFFLES~\citep{raffles}, DeepDebug~\citep{agentdebugx}, DCFA~\citep{dcfa}, and CHIEF~\citep{chief}. We rerun all baselines under a common setup. Appendix~\ref{sec:appendix-baselines} provides implementation and adaptation details.

\paragraph{Metrics.}
Our primary metric is exact-match accuracy for the decisive error step, corresponding to Overall Step on Who\&When and Root Exact on LongRCA. We refer to both as exact step-localization accuracy. We also report responsible-agent accuracy (Agent on Who\&When and Role on LongRCA), as well as the proportion of predictions within five steps of the annotated error step (Root \(\pm\)5) and the mean absolute error between predicted and annotated step indices (Root MAE) on LongRCA. We measure cost by the average token consumption per trace (Tok.), counting input and output tokens across all model calls within one execution of the method. We use Cost-of-Pass (CoP)~\citep{costofpass}, measured in tokens, to jointly assess localization accuracy and token cost. It is defined as total token consumption divided by the number of correctly localized traces. Scoring details, token accounting, and token breakdowns are provided in Appendices~\ref{sec:appendix-scoring} and~\ref{sec:token-breakdown}.

\paragraph{Backbones and implementation.}
The main experiments use DeepSeek-V4.1-Flash~\citep{deepseekv41flash} and GLM-5.3-Flash~\citep{glm5} as backbone models, yielding four evaluation settings across the two benchmarks. We additionally use Claude Sonnet 5.5~\citep{sonnet55} to compare CoLFA with five baselines on all failed execution traces from Who\&When. These baselines cover both attribution strategies. CoLFA uses the same rules and parameters to identify execution links on both benchmarks. CoLFA's attribution input and prompts are provided in Appendices~\ref{sec:appendix-input} and~\ref{sec:appendix-prompts}. Generation settings for all methods are provided in Appendix~\ref{sec:appendix-baselines}.

\subsection{Main Results}

\label{sec:exp-main}

\begin{table*}[t]
  \centering
  \footnotesize
  \setlength{\tabcolsep}{0pt}
  \renewcommand{\arraystretch}{1.0}
  \setlength{\aboverulesep}{0.3ex}
  \setlength{\belowrulesep}{0.3ex}
  \setlength{\cmidrulesep}{0.2ex}
  \ifdefined\accw\else\newlength{\accw}\newlength{\copw}\fi
  \settowidth{\accw}{00.00}\settowidth{\copw}{000}
  \begin{tabular*}{\textwidth}{@{\extracolsep{\fill}}l@{\hspace{4pt}}rrrrrrrr@{\hspace{6pt}}rrrrrrrr@{}}
    \toprule
    & \multicolumn{8}{c}{\textbf{DeepSeek-V4.1-Flash}} & \multicolumn{8}{c}{\textbf{GLM-5.3-Flash}} \\
    \cmidrule(lr){2-9} \cmidrule(lr){10-17}
    & \multicolumn{2}{c}{Alg.} & \multicolumn{2}{c}{Hand.} & \multicolumn{2}{c}{Overall} & \multicolumn{2}{c}{Cost (K)} & \multicolumn{2}{c}{Alg.} & \multicolumn{2}{c}{Hand.} & \multicolumn{2}{c}{Overall} & \multicolumn{2}{c}{Cost (K)} \\
    \cmidrule(lr){2-3} \cmidrule(lr){4-5} \cmidrule(lr){6-7} \cmidrule(lr){8-9} \cmidrule(lr){10-11} \cmidrule(lr){12-13} \cmidrule(lr){14-15} \cmidrule(lr){16-17}
    Method & \makebox[\accw][c]{Agent} & \makebox[\accw][c]{Step} & \makebox[\accw][c]{Agent} & \makebox[\accw][c]{Step} & \makebox[\accw][c]{Agent} & \makebox[\accw][c]{Step} & \makebox[\accw][c]{Tok.} & \makebox[\copw][r]{CoP} & \makebox[\accw][c]{Agent} & \makebox[\accw][c]{Step} & \makebox[\accw][c]{Agent} & \makebox[\accw][c]{Step} & \makebox[\accw][c]{Agent} & \makebox[\accw][c]{Step} & \makebox[\accw][c]{Tok.} & \makebox[\copw][r]{CoP} \\
    \midrule
    All-at-Once & 60.32 & 23.81 & \underline{60.34} & 3.45 & 60.33 & 17.39 & 8.4 & 48 & 65.08 & 12.70 & 53.45 & 3.45 & 61.41 & 9.78 & \underline{8.2} & 84 \\
    A2P & \underline{62.70} & 45.24 & 56.90 & 20.69 & \underline{60.87} & 37.50 & \underline{8.3} & \underline{22} & 64.29 & 50.00 & \underline{56.90} & 20.69 & 61.96 & 40.76 & 8.4 & \underline{21} \\
    CORRECT & 61.11 & \underline{50.00} & 48.28 & 22.41 & 57.07 & \underline{41.30} & 22.7 & 55 & \underline{66.67} & \underline{53.17} & 48.28 & \underline{25.86} & 60.87 & \underline{44.57} & 20.6 & 46 \\
    Step-by-Step & 42.06 & 32.54 & 51.72 & 20.69 & 45.11 & 28.80 & 111.1 & 386 & 46.03 & 35.71 & 50.00 & 18.97 & 47.28 & 30.43 & 32.0 & 105 \\
    Binary Search & 32.54 & 4.76 & \underline{60.34} & 12.07 & 41.30 & 7.07 & 14.7 & 208 & 54.76 & 36.51 & 44.83 & 10.34 & 51.63 & 28.26 & 15.3 & 54 \\
    ECHO & 61.90 & 48.41 & 44.83 & 22.41 & 56.52 & 40.22 & 60.9 & 151 & 54.76 & \underline{53.17} & 53.45 & 24.14 & 54.35 & 44.02 & 58.5 & 133 \\
    RAFFLES & 61.90 & 46.83 & 56.90 & 20.69 & 60.33 & 38.59 & 34.6 & 90 & \underline{66.67} & 50.79 & 55.17 & 20.69 & \underline{63.04} & 41.30 & 34.2 & 83 \\
    DeepDebug & 48.41 & 37.30 & \underline{60.34} & 18.97 & 52.17 & 31.52 & 13.7 & 44 & 57.94 & 44.44 & 53.45 & 20.69 & 56.52 & 36.96 & 14.2 & 38 \\
    DCFA & 52.38 & 40.48 & \underline{60.34} & \textbf{31.03} & 54.89 & 37.50 & 107.1 & 285 & 59.52 & 43.65 & \underline{56.90} & 20.69 & 58.70 & 36.41 & 141.2 & 388 \\
    CHIEF & 60.32 & 36.51 & 55.17 & 13.79 & 58.70 & 29.35 & 72.9 & 248 & 64.29 & 41.27 & \underline{56.90} & 10.34 & 61.96 & 31.52 & 69.3 & 220 \\
    \specialrule{\lightrulewidth}{\aboverulesep}{0pt}
    \shaderow{colfarow}
    \shadestrut \textbf{\textcolor{colfatext}{\method}} & \textbf{65.87} & \textbf{52.38} & \textbf{62.07} & \underline{27.59} & \textbf{64.67} & \textbf{44.57} & \textbf{6.9} & \textbf{16} & \textbf{68.25} & \textbf{54.76} & \textbf{62.07} & \textbf{29.31} & \textbf{66.30} & \textbf{46.74} & \textbf{7.1} & \textbf{15} \\
    \specialrule{\heavyrulewidth}{0pt}{\belowbottomsep}
  \end{tabular*}
  \caption{Main results on Who\&When with two backbones. Bold and underline mark the best and second best.}
  \label{tab:ww}
\end{table*}

\begin{table*}[t]
  \centering
  \footnotesize
  \setlength{\tabcolsep}{0pt}
  \renewcommand{\arraystretch}{1.0}
  \setlength{\aboverulesep}{0.3ex}
  \setlength{\belowrulesep}{0.3ex}
  \setlength{\cmidrulesep}{0.2ex}
  \ifdefined\lraccw\else\newlength{\lraccw}\newlength{\lrcopw}\fi
  \settowidth{\lraccw}{00.00}\settowidth{\lrcopw}{12{,}449}
  \begin{tabular*}{\textwidth}{@{\extracolsep{\fill}}l@{\hspace{4pt}}rrrrrr@{\hspace{6pt}}rrrrrr@{\hspace{2.4pt}}}
    \toprule
    & \multicolumn{6}{c}{\textbf{DeepSeek-V4.1-Flash}} & \multicolumn{6}{c}{\textbf{GLM-5.3-Flash}} \\
    \cmidrule(lr){2-7} \cmidrule(lr){8-13}
    Method & \makebox[\lraccw][c]{Role} & \makebox[\lraccw][c]{Exact} & \makebox[\lraccw][c]{$\pm$5} & \makebox[\lraccw][c]{MAE} & \makebox[\lraccw][c]{Tok.\,(K)} & \makebox[\lrcopw][c]{CoP\,(K)} & \makebox[\lraccw][c]{Role} & \makebox[\lraccw][c]{Exact} & \makebox[\lraccw][c]{$\pm$5} & \makebox[\lraccw][c]{MAE} & \makebox[\lraccw][c]{Tok.\,(K)} & \makebox[\lrcopw][c]{CoP\,(K)} \\
    \midrule
    All-at-Once & 42.02 & 1.40 & 10.79 & 77.09 & \underline{52.8} & 3{,}763 & 37.81 & 1.05 & 11.84 & 60.06 & \underline{52.3} & 4{,}969 \\
    A2P & 46.49 & 13.51 & 29.47 & 43.07 & 53.6 & \underline{397} & 47.02 & 16.49 & 28.16 & 43.65 & 52.9 & \underline{321} \\
    CORRECT & \underline{49.65} & \underline{18.86} & 25.70 & 41.88 & 114.5 & 607 & 49.21 & 20.35 & 32.19 & 40.04 & 110.3 & 542 \\
    Step-by-Step & 33.42 & 7.63 & 17.37 & 46.16 & 950.0 & 12{,}449 & 30.61 & 8.33 & 16.67 & 54.62 & 367.8 & 4{,}414 \\
    Binary Search & 15.96 & 0.00 & 15.96 & 77.79 & 130.4 & --- & 28.51 & 3.51 & 18.07 & 53.47 & 114.7 & 3{,}268 \\
    ECHO & 44.47 & 15.88 & 28.51 & 48.94 & 347.4 & 2{,}188 & 41.67 & 16.67 & 28.51 & 47.15 & 337.6 & 2{,}025 \\
    RAFFLES & 48.60 & 18.33 & \underline{29.82} & \underline{40.59} & 215.9 & 1{,}178 & \textbf{54.21} & \underline{21.75} & \underline{34.74} & \underline{35.97} & 212.3 & 976 \\
    DeepDebug & 47.19 & 17.63 & 26.40 & 52.08 & 71.3 & 404 & 38.16 & 17.11 & 26.40 & 46.06 & 71.9 & 421 \\
    \specialrule{\lightrulewidth}{\aboverulesep}{0pt}
    \shaderow{colfarow}
    \shadestrut \textbf{\textcolor{colfatext}{\method}} & \textbf{51.58} & \textbf{21.84} & \textbf{32.81} & \textbf{35.54} & \textbf{22.4} & \textbf{103} & \underline{52.28} & \textbf{23.60} & \textbf{36.23} & \textbf{35.42} & \textbf{22.0} & \textbf{93} \\
    \specialrule{\heavyrulewidth}{0pt}{\belowbottomsep}
  \end{tabular*}
  \caption{Main results on LongRCA with two backbones. Bold and underline mark the best and second best.}
  \label{tab:lr}
\end{table*}

CoLFA achieves the highest exact step-localization accuracy and lowest token cost and CoP in all four settings, exceeding each setting's most accurate baseline by 1.84--3.26\% with 66--90\% fewer tokens. On LongRCA, which has longer traces on average, trace-level paired bootstrap further supports CoLFA's exact step-localization advantage. The 95\% confidence intervals for accuracy differences are $[0.44, 5.53]$\% and $[0.28, 3.51]$\% for DeepSeek and GLM, respectively (Appendix~\ref{sec:appendix-bootstrap}).

Table~\ref{tab:ww} shows that on Who\&When, CoLFA achieves the highest exact step-localization accuracy with the lowest token consumption under both backbones, reducing CoP by 27\% to 30\% relative to the lowest baseline CoP in each setting. Across comparisons with individual baselines under the two backbones, CoLFA's largest step-localization gains are 47.62\% and 25.86\% on Alg. and Hand., respectively. Team composition and division of work vary by task in Alg., whereas Hand. involves more orchestration and subsequent processing. Both types of systems require linking executions across agents and distinguishing the roles of different processing steps. CoLFA uses execution links and shared points of comparison to support comparisons between earlier and later processing, retains each decision's context, and reduces the processing of lengthy histories and repeated text.

As shown in Table~\ref{tab:lr}, on LongRCA, CoLFA achieves the highest Root Exact with the lowest token consumption under both backbones, reducing CoP by 71\% to 74\% relative to the lowest baseline CoP in each setting. Across comparisons with individual baselines under the two backbones, its largest Root Exact gain is 22.54\%. CoLFA also achieves the highest Root \(\pm\)5 and the lowest Root MAE under both backbones. CoLFA combines detailed records of the selected steps with a trace overview covering the entire execution. This organization allows the model to check processing at individual steps against subsequent execution progress when assessing the role of errors in the final failure, while reducing the amount of history presented in detail and the repetition in the input.

\begin{figure}[t]
  \centering
  \includegraphics[width=\columnwidth]{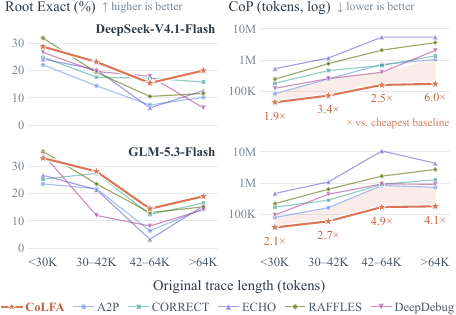}
  \caption{Root Exact and CoP on LongRCA across four equal-sized groups by original trace token count.}
  \label{fig:lr-length}
\end{figure}

To compare localization performance and cost across trace lengths, we divide LongRCA into four equal-sized groups by original trace token count. As shown in Figure~\ref{fig:lr-length}, under both backbones, CoLFA achieves the lowest CoP in every length group and the highest Root Exact in the longest group, where its CoP is 75\% to 83\% lower than the lowest baseline CoP. These results show that CoLFA's advantage in cost per correct localization holds across trace lengths, with joint advantages in localization accuracy and cost in the longest group.

\begin{table}[t]
  \centering
  \footnotesize
  \setlength{\tabcolsep}{0pt}
  \renewcommand{\arraystretch}{1.0}
  \setlength{\aboverulesep}{0.3ex}
  \setlength{\belowrulesep}{0.3ex}
  \setlength{\cmidrulesep}{0.2ex}
  \ifdefined\saccw\else\newlength{\saccw}\newlength{\stokw}\newlength{\scopw}\fi
  \settowidth{\saccw}{00.00}\settowidth{\stokw}{00.0}\settowidth{\scopw}{000}
  \resizebox{\columnwidth}{!}{\begin{tabular}{@{}l@{\hspace{8pt}}r@{\hspace{4pt}}r@{\hspace{4pt}}r@{\hspace{4pt}}r@{\hspace{4pt}}r@{\hspace{4pt}}r@{\hspace{4pt}}r@{\hspace{4pt}}r@{}}
    \toprule
    & \multicolumn{2}{c}{Alg.} & \multicolumn{2}{c}{Hand.} & \multicolumn{2}{c}{Overall} & \multicolumn{2}{c}{Cost (K)} \\
    \cmidrule(lr){2-3} \cmidrule(lr){4-5} \cmidrule(lr){6-7} \cmidrule(lr){8-9}
    Method & \makebox[\saccw][c]{Agent} & \makebox[\saccw][c]{Step} & \makebox[\saccw][c]{Agent} & \makebox[\saccw][c]{Step} & \makebox[\saccw][c]{Agent} & \makebox[\saccw][c]{Step} & \makebox[\stokw][c]{Tok.} & \makebox[\scopw][c]{CoP} \\
    \midrule
    All-at-Once & \underline{67.46} & 39.68 & 62.07 & 3.45 & 65.76 & 28.26 & \underline{12.4} & 44 \\
    A2P & 65.87 & \underline{50.00} & \textbf{67.24} & 29.31 & \underline{66.30} & 43.48 & 13.3 & \underline{31} \\
    CORRECT & 65.87 & 47.62 & \textbf{67.24} & \underline{32.76} & \underline{66.30} & 42.93 & 33.6 & 78 \\
    RAFFLES & 65.08 & 47.62 & 62.07 & \textbf{36.21} & 64.13 & \underline{44.02} & 53.5 & 122 \\
    DeepDebug & 62.70 & 47.62 & 44.83 & 20.69 & 57.07 & 39.13 & 21.0 & 54 \\
    \specialrule{\lightrulewidth}{\aboverulesep}{0pt}
    \shaderow{colfarow}
    \shadestrut \textbf{\textcolor{colfatext}{\method}} & \textbf{71.43} & \textbf{52.38} & \underline{65.52} & \underline{32.76} & \textbf{69.57} & \textbf{46.20} & \textbf{10.7} & \textbf{23} \\
    \specialrule{\heavyrulewidth}{0pt}{\belowbottomsep}
  \end{tabular}}
  \caption{Results with Claude Sonnet 5.5 on Who\&When under strict matching.}
  \label{tab:sonnet-ww}
\end{table}

To test whether this joint advantage also holds with a frontier model, we additionally compare CoLFA with five baselines on Who\&When using Claude Sonnet 5.5. As shown in Table~\ref{tab:sonnet-ww}, CoLFA again achieves the highest exact step-localization accuracy with the lowest token consumption, reducing CoP by 25\% relative to the lowest baseline CoP. CoLFA organizes the attribution input according to how results are formed through multi-agent collaboration and what failure attribution needs to check. Because these principles do not change with the attribution backbone, this organization can help different models link executions and compare earlier and later processing while reducing the processing of lengthy and repeated records.

\subsection{Ablation Study}

\label{sec:exp-ablation}

\begin{table*}[t]
  \centering
  \footnotesize
  \setlength{\tabcolsep}{0pt}
  \renewcommand{\arraystretch}{1.0}
  \setlength{\aboverulesep}{0.3ex}
  \setlength{\belowrulesep}{0.3ex}
  \setlength{\cmidrulesep}{0.2ex}
  \ifdefined\ablaccw\else\newlength{\ablaccw}\newlength{\ablrelw}\fi
  \settowidth{\ablaccw}{00.00}\settowidth{\ablrelw}{0.00}
  \begin{tabular*}{\textwidth}{@{\extracolsep{\fill}}l@{\hspace{4pt}}rrrr@{\hspace{6pt}}rrrr@{\hspace{6pt}}rrrr@{\hspace{6pt}}rrrr@{}}
    \toprule
    & \multicolumn{4}{c}{\makebox[0pt][c]{Who\&When / DeepSeek}} & \multicolumn{4}{c}{\makebox[0pt][c]{Who\&When / GLM}} & \multicolumn{4}{c}{\makebox[0pt][c]{LongRCA / DeepSeek}} & \multicolumn{4}{c}{\makebox[0pt][c]{LongRCA / GLM}} \\
    \cmidrule(lr){2-5} \cmidrule(lr){6-9} \cmidrule(lr){10-13} \cmidrule(lr){14-17}
    Variant & \makebox[\ablaccw][c]{Agent} & \makebox[\ablaccw][c]{Step} & \makebox[\ablrelw][c]{Tok.} & \makebox[\ablrelw][c]{CoP} & \makebox[\ablaccw][c]{Agent} & \makebox[\ablaccw][c]{Step} & \makebox[\ablrelw][c]{Tok.} & \makebox[\ablrelw][c]{CoP} & \makebox[\ablaccw][c]{Role} & \makebox[\ablaccw][c]{Exact} & \makebox[\ablrelw][c]{Tok.} & \makebox[\ablrelw][c]{CoP} & \makebox[\ablaccw][c]{Role} & \makebox[\ablaccw][c]{Exact} & \makebox[\ablrelw][c]{Tok.} & \makebox[\ablrelw][r]{CoP} \\
    \specialrule{\lightrulewidth}{\aboverulesep}{0pt}
    \shaderow{colfarow}
    \shadestrut \textbf{\textcolor{colfatext}{\method}} & \textbf{64.67} & \textbf{44.57} & \textbf{1.00} & \textbf{1.00} & \textbf{66.30} & \textbf{46.74} & \textbf{1.00} & \textbf{1.00} & \textbf{51.58} & \textbf{21.84} & \textbf{1.00} & \textbf{1.00} & \textbf{52.28} & \textbf{23.60} & \textbf{1.00} & \textbf{1.00} \\
    w/o Expansion & 59.78 & 39.13 & 1.10 & 1.25 & 61.96 & 42.39 & 1.09 & 1.21 & 47.02 & 19.65 & 1.47 & 1.63 & 49.82 & 20.70 & 1.47 & 1.68 \\
    w/o Organization & 60.87 & 40.22 & 1.09 & 1.20 & 62.50 & 43.48 & 1.08 & 1.17 & 49.47 & 19.30 & 1.35 & 1.52 & 48.07 & 20.35 & 1.36 & 1.57 \\
    w/o Both & 57.07 & 36.96 & 1.21 & 1.46 & 57.61 & 39.13 & 1.21 & 1.44 & 45.26 & 17.89 & 2.32 & 2.83 & 46.32 & 19.30 & 2.32 & 2.83 \\
    \bottomrule
  \end{tabular*}
  \caption{Ablation study of \method. Tok. and CoP are relative to \method. Bold marks the best values.}
  \label{tab:ablation}
\end{table*}

To assess the contributions of step-level scope organization and content-level presentation to localization accuracy and token cost, we compare CoLFA with three ablation variants across the four main settings. w/o Expansion retains content organization and includes all decision steps in the detailed records and the set of candidate steps for the model's answer. w/o Organization keeps the formation steps and answer candidates unchanged, presents each step's complete original text without truncation, and removes content references, step references, and the trace overview. w/o Both gives the model the complete original text of all decision steps and allows it to select from all of them, serving as an internal full-trace control. All variants use the same attribution prompt as CoLFA. Detailed settings are provided in Appendix~\ref{sec:appendix-ablation}.

Table~\ref{tab:ablation} shows that removing either component lowers exact step-localization accuracy and increases token consumption. Both components thus contribute to joint improvements in accuracy and token cost by organizing the evidence available to the model in different ways. w/o Expansion includes all decision steps in the detailed records and answer candidates, yet accuracy decreases and token consumption increases. This highlights the value of linking executions and setting expansion scopes according to what needs to be checked when addressing Interleaved Steps. w/o Organization keeps the formation steps and answer candidates unchanged, yet both metrics worsen. This shows that, after selecting steps for examination, their evidence still needs to be organized to address Cluttered Content, allowing earlier and later processing to be compared and the role of errors to be assessed in light of overall execution progress. w/o Both has lower accuracy and higher token consumption than either single-component ablation under both backbones on both benchmarks, showing the consistency of the joint contribution of both levels of organization across trace conditions and attribution backbones.

\subsection{Analysis of Attribution Inputs}

\label{sec:exp-analysis}

\begin{figure}[t]
  \centering
  \includegraphics[width=\columnwidth]{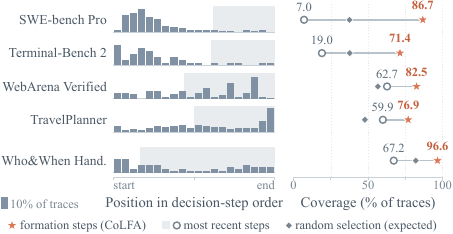}
  \caption{Decisive error position distributions (left) and coverage under matched step counts (right) on multi-agent LongRCA traces and Who\&When Hand.}
  \label{fig:error-position}
\end{figure}

To examine whether collaboration cues provide useful evidence for attribution, we analyze CoLFA's attribution inputs. We examine coverage of decisive error steps, and the textual basis for shared points of comparison and input size. Detailed statistical procedures, experimental settings, and complete supplementary results are provided in Appendix~\ref{sec:appendix-analysis}.

To test whether collaboration cues help retain decisive error steps while controlling expansion size, we measure coverage on multi-agent traces from LongRCA and the Hand. split of Who\&When. Coverage is the proportion of traces whose formation steps contain the annotated decisive error step. We compare against recency and random selection, matching CoLFA's number of decision steps for each trace. Figure~\ref{fig:error-position} shows higher coverage for collaboration-guided selection than both alternatives. On LongRCA's multi-agent traces, coverage exceeds recency selection by 26.7\%. A supplementary matched-step attribution experiment on LongRCA also yields higher localization accuracy than both alternatives under both backbones (Appendix~\ref{sec:appendix-step-selection}). These results show that execution links left as multi-agent systems form results can guide step selection for failure attribution.

To examine the textual basis for shared points of comparison and organized input size, we analyze content references and input size on the same multi-agent LongRCA traces. Content references appear in almost every trace, replacing repeated spans that account for 24\% of the pre-truncation characters in actions and tool outputs or environment feedback within detailed records. After organizing steps and content, the aggregate token count of attribution inputs is 40\% of that of complete traces. The widespread presence of shared text provides reference points for comparing how different agents process the same content. While reducing overall input size, CoLFA shares original text through references and retains a separate record for each execution, allowing the model to compare earlier and later processing and assess each decision in its own context.

\section{Conclusion}
\label{sec:conclusion}

We propose \method, which uses collaboration cues from result formation to organize which steps are examined and how their content is presented for multi-agent failure attribution.
Across the evaluated settings on Who\&When and LongRCA, \method achieves the highest exact step-localization accuracy with the lowest token consumption.
It offers more accurate diagnostics at lower cost for agent debugging and iterative improvement.

\section*{Limitations}

\method uses observable collaboration cues in execution traces to organize which steps are examined for failure attribution in multi-agent systems. Our current expansion scopes are defined for collaboration configurations involving delegation and handoffs or broadcast group chats. The resulting formation steps define the scope of detailed records and the candidate set for decisive error localization. This scope depends on whether recorded handoffs and content correspondence link the relevant executions. The applicability of execution links and expansion scopes to collaboration configurations and record formats not evaluated in this study requires further investigation.

We evaluate decisive error localization and token consumption on Who\&When and LongRCA, with attribution judgments based on recorded executions. Whether attribution results support effective repair and improve task completion remains to be directly validated by correcting the relevant steps and examining the outcomes of subsequent reruns.

\bibliography{references}

\clearpage
\appendix
\makeatletter
\setlength{\@fptop}{0pt}
\setlength{\@dblfptop}{0pt}
\setlength{\@dblfpsep}{12pt}
\setlength{\@dblfpbot}{0pt plus 1fil}
\renewcommand{\dbltopfraction}{0.85}
\makeatother

\section{Execution Links and Expansion Scope}
\label{sec:appendix-links}

\subsection{Execution Link Rules}
\label{sec:appendix-link-rules}

Multi-agent systems connect executions by different agents through task assignments, content handoffs, and subsequent processing. Assessing whether the processing at a step is erroneous requires examining its task requirements and inputs, which may be recorded in earlier steps by other agents. Handoff records capture the transfer of requirements and intermediate results, while content correspondence across steps suggests that different executions may involve the same information. We use the following rules to identify earlier steps that can be revisited when examining step \(i\). Their original indices form \(P(i)\), which represents the execution links of step \(i\). These links provide pointers for examining related executions together. The model uses each execution's task requirements, information available at that step, actual actions, and subsequent progress to assess how content was used and whether an error occurred and was decisive.

\paragraph{Handoffs.}
Assignment records specify what work the receiving agent should perform, so checking its processing requires revisiting these requirements. Systems that delegate work typically identify the receiving agent in the author field or step content, for example with \texttt{(-> X)} or a \texttt{delegate\_to\_X} call. For a decision step by agent X, we first link to the most recent earlier step in which another agent assigned work to X. If no explicit assignment is recorded, we link to the most recent earlier non-tool step by another agent that mentions X, since naming a recipient in text may also convey requirements or hand off content. When adjacent decision steps have different authors, we also link the later step to the earlier one, using the change in speaking agent as an additional cue for connecting executions. These links allow the receiving agent's actions to be checked against what it was asked to do, helping distinguish erroneous requirements from errors in interpreting or carrying them out.

\paragraph{Content correspondence.}
The same information may recur in records of content production, handoffs, and subsequent processing. Linking these records helps check how the information was formed and later processed. We find related executions using numerical values, identifiers, and specific text that recur across steps, detecting numbers, identifiers, and quoted text that meet their respective minimum length requirements. We exclude values already given in the task description, agent names, and common content because they do not readily distinguish which specific records should be revisited. We measure an item's specificity by its inverse document frequency across the original records of a single trace. We retain only items with specificity at least \(\theta=1.4\), corresponding to occurrence in no more than approximately one quarter of the records. The resulting links focus on distinctive content. Threshold sensitivity is examined in Table~\ref{tab:theta}.

For each retained item in step \(i\), we collect its earlier occurrences and retain the first occurrence for each record author as a candidate. These candidates allow the model to revisit how the content was handled when it first entered each participant's records. Reviewing, reporting, and coordinating roles may restate content produced at other steps. We therefore prefer the most recent candidate that is not a reporting step. If no such candidate is available, we select the most recent reporting step. We identify reporting steps from role names and wording that reports, summarizes, or closes the task. This selection directs links toward earlier production or processing and reduces links to later restatements. It helps the model check whether an error had already occurred in an earlier execution and whether later executions misread, modified, or corrected the content.

Tool outputs present the results of actions. Attribution also requires examining the agent decisions that obtained or processed this content. If the selected occurrence is a tool-output record, we search the item's earlier candidates for the most recent agent step. If none is available, we link to the step that issued the tool call. This makes the linked record a decision step that can be assessed for attribution. Each item contributes at most one earlier step. Links from different items are combined with handoff links in \(P(i)\).

\paragraph{Parameter settings.}
Both benchmarks use a specificity threshold of \(\theta=1.4\) for content correspondence. Specificity is computed from record frequencies within each trace, so the same threshold applies a common relative-frequency criterion across tasks to select distinctive content correspondence cues. Table~\ref{tab:params} also lists the parameters for content references and truncation of detailed records, whose uses are described in Appendix~\ref{sec:appendix-input}. The content-reference parameters are the same on both benchmarks. Given the longer execution traces in LongRCA, we use shorter truncation lengths for its detailed records to control the amount of lengthy reasoning and tool outputs or environment feedback presented to the model.

\begin{table}[!htbp]
  \centering
  \footnotesize
  \setlength{\tabcolsep}{0pt}
  \renewcommand{\arraystretch}{1.0}
  \setlength{\aboverulesep}{0.3ex}
  \setlength{\belowrulesep}{0.3ex}
  \setlength{\cmidrulesep}{0.2ex}
  \begin{tabular*}{\columnwidth}{@{\extracolsep{\fill}}>{\raggedright\arraybackslash}p{4.3cm} N{6{,}000} N{6{,}000}@{}}
    \toprule
    Parameter & \hc{Who\&When} & \hc{LongRCA} \\
    \midrule
    \multicolumn{3}{@{}l}{\textit{Execution links and content references}} \\
    Specificity threshold $\theta$ & \multicolumn{2}{c@{}}{1.4} \\
    Minimum length of a region replaced by a content reference (words) & \multicolumn{2}{c@{}}{12} \\
    \midrule
    \multicolumn{3}{@{}l}{\textit{Detailed records (characters)}} \\
    Reasoning truncation length & 1{,}500 & 500 \\
    Tool output or environment feedback truncation length & 6{,}000 & 1{,}500 \\
    \bottomrule
  \end{tabular*}
  \caption{CoLFA parameters. Truncation lengths for detailed records are measured in characters.}
  \label{tab:params}
\end{table}

\subsection{Expansion Scope}
\label{sec:appendix-scope}

Execution links indicate which earlier executions to revisit, while the expansion scope determines which steps and how much history to include along these links. Tracing how content was produced requires examining relevant computations, judgments, and actions. Checking subsequent processing requires the recipient's task requirements, available information, and actual actions. Requirements and inputs supplied by other participants provide additional evidence for these checks. CoLFA accordingly sets how much history to include for each agent and constructs these scopes according to the system's known collaboration configuration. Differences in scope reflect what each record helps check, while agent identities are used to apply the selection.

\paragraph{Expansion scope and collaboration configuration.}
In systems where agents delegate and hand off work, we construct output steps, production steps, and supply steps as defined in Section~\ref{sec:method-expansion}. The output agent may receive and adopt information, take actions, and make revisions multiple times before producing the final result. We therefore retain all of its decision steps throughout the trace. This allows its processing to be examined in sequence, supporting checks of whether it introduced an error and how it handled earlier content. The retained steps also provide multiple entry points for tracing related executions.

Handed-off content presents the outcome of upstream work. Checking whether that outcome is erroneous also requires revisiting the computations, judgments, or actions that produced it. We select as the upstream agent the author of the last decision step not taken by the output agent. We obtain entry points within this agent from the execution links of all output steps, then expand only within that agent using content correspondence to obtain the production steps. These steps allow processing at the receiving end to be checked against the earlier basis for producing the content. This helps distinguish errors already introduced upstream from downstream errors in interpretation or processing, while limiting the upstream history included in detailed examination to steps connected through these links.

Assessing the processing in output and production steps also depends on the requirements and inputs supplied by the remaining agents. Supply steps therefore include earlier steps by these agents that are directly linked to either set, and expansion stops there. These records provide a basis for checking what requirements the processing should satisfy and what inputs informed the decisions. Supply steps belong to both the formation steps and the set of steps from which the model may answer, allowing the model to check whether the requirements or inputs themselves are erroneous. Stopping here limits further expansion into the other participants' internal histories.

A broadcast group chat shares messages with the entire team, so the output agent may form its result using any message from the upstream agent. In this configuration, we include all decision steps of the upstream agent as production steps, retaining the scope needed to examine the content it supplied. The remaining agents are still included through supply steps. A single-agent execution has no upstream agent. Its formation steps are the output steps, \(S=S_o\), and the full sequence of its decision steps shows how the result was formed.

In our evaluation, Magentic-One in Who\&When Hand-Crafted and the multi-agent systems in LongRCA use the scope for delegation and handoffs. CaptainAgent in Who\&When Algorithm-Generated records no assignments and shares all messages with the group chat, so it uses the broadcast group chat scope. All 108 VitaBench traces in LongRCA are single-agent executions.

\paragraph{Scope comparison and analysis.}
To examine the trade-off between coverage and size under different expansion scopes, we compare the five scopes in Table~\ref{tab:scope} on the 1,032 multi-agent traces from LongRCA without model calls. Coverage is the proportion of traces whose selected steps contain the annotated decisive error step. The mean step share is the average of \(|S|/|V|\), calculated separately for each trace. Each strategy supplements its selected steps with directly linked earlier steps, so this additional set varies with the expansion scope.

\begin{table}[!htbp]
  \centering
  \footnotesize
  \setlength{\tabcolsep}{0pt}
  \renewcommand{\arraystretch}{1.0}
  \setlength{\aboverulesep}{0.3ex}
  \setlength{\belowrulesep}{0.3ex}
  \setlength{\cmidrulesep}{0.2ex}
  \begin{tabular*}{\columnwidth}{@{\extracolsep{\fill}}>{\raggedright\arraybackslash}p{5.0cm} N{00.0} N{00.0}@{}}
    \toprule
    Expansion scope & \hc{Coverage} & \hc{\shortstack{Mean step\\share}} \\
    \midrule
    All decision steps of the output and upstream agents, supplemented with directly linked steps & 87.4 & 71.9 \\
    \shadelines{colfarow}{2}
    \shadestrutht Output steps, upstream production steps, and their supply steps \textcolor{colfatext}{(\method)}\shadestrutend & 78.9 & 49.3 \\
    All decision steps of the output agent, supplemented with directly linked steps & 67.2 & 40.7 \\
    All decision steps of the upstream agent, backward tracing within the output agent through content correspondence, and directly linked steps & 61.3 & 52.3 \\
    Backward tracing within both the output and upstream agents through content correspondence and assignments, supplemented with directly linked steps & 47.5 & 26.1 \\
    \bottomrule
  \end{tabular*}
  \caption{Comparison of expansion scopes on multi-agent traces from LongRCA. Values are percentages. Mean step share is the mean of \(|S|/|V|\) across traces.}
  \label{tab:scope}
\end{table}

CoLFA has a mean step share of 49.3\% and coverage of 78.9\%. Retaining all decision steps of the upstream agent and tracing backward within the output agent through content correspondence increases the mean step share to 52.3\% but lowers coverage to 61.3\%. When backward tracing within both agents uses only content correspondence and assignments, the mean step share is 26.1\% and coverage falls to 47.5\%. These results support using different scopes for the two agents. Retaining all decision steps of the output agent and tracing upstream production through execution links allows the formation of the final result and the earlier processing of relevant content to be examined together.

Starting from all decision steps of the output agent and their directly linked records, further tracing upstream production and adding the corresponding supply steps increases coverage from 67.2\% to 78.9\% and the mean step share from 40.7\% to 49.3\%. This additional scope reaches earlier steps that produced the handed-off content and incorporates other requirements and inputs needed to assess those steps. Retaining all decision steps of both the output and upstream agents further raises coverage to 87.4\%, with a mean step share of 71.9\%. Of the 20,126 upstream steps excluded by CoLFA, 76 are annotated decisive error steps. These results characterize the trade-off in the current scope. Tracing upstream production expands error coverage while controlling how much internal upstream history is included.

Stopping expansion at supply steps adds directly relevant requirements and inputs while limiting further inclusion of other participants' internal histories. In the closure comparison, starting from the final step and repeatedly following both execution links and each agent's preceding steps expands the scope to all decision steps. This shows why expansion paths and stopping points need to be specified. These analyses measure how well different scopes retain annotated error steps and characterize their trade-offs in size. Final localization performance is evaluated in the main experiments and component ablations.

\section{Attribution Input}
\label{sec:appendix-input}

After step expansion identifies the formation steps \(S\) and execution links \(P(i)\), attribution still requires examining the processing in these executions and assessing the role of errors in light of overall execution progress. Following Section~\ref{sec:method-organization}, CoLFA retains a separate detailed record for each formation step, uses content references and step references to support comparisons between earlier and later processing, and supplements these records with a trace overview of progress throughout the execution. The task description, detailed records, and trace overview form the attribution input. The model uses this input together with the outcome information and judgment criteria in the prompt to identify the decisive error step.

Algorithm~\ref{alg:colfa} summarizes the complete procedure from step expansion to the final judgment. Here, \(P(U)\) denotes the union of the execution links of the steps indexed by \(U\), and \(\mathrm{cl}(\cdot)\) denotes closure under content correspondence within the selected upstream agent only. \(V_h\) is the set of decision step indices of that upstream agent, and \(V_p\) is the set of decision step indices of agents other than the output agent and the selected upstream agent. Appendix~\ref{sec:appendix-links} describes the execution links and expansion scopes for different collaboration configurations. Record organization and judgment prompts are described below.

\begin{algorithm}[!htbp]
  \caption{CoLFA}
  \label{alg:colfa}
  \small
  \algrenewcommand\algorithmicrequire{\textbf{Input.}}
  \algrenewcommand\algorithmicensure{\textbf{Output.}}
  \begin{algorithmic}[1]
    \Require A task description \(q\), a failed execution trace \(\tau\), and a reference answer if provided by the evaluation protocol, as described in Appendix~\ref{sec:appendix-benchmarks}.
    \Ensure The predicted decisive error step \(\hat{s}\) and responsible agent.
    \State Obtain the execution links \(P(i)\) of each decision step from handoffs and content correspondence, following Appendix~\ref{sec:appendix-links}.
    \State Identify the output agent and selected upstream agent as defined in Section~\ref{sec:method-expansion}. Set \(S_o\leftarrow\) all decision steps of the output agent.
    \State If there is no upstream agent, set \(S_h\leftarrow\emptyset\). For a broadcast group chat, set \(S_h\leftarrow V_h\). For delegation and handoffs, set \(S_h\leftarrow\mathrm{cl}(P(S_o)\cap V_h)\).
    \State Set \(S_p\leftarrow P(S_o\cup S_h)\cap V_p\).
    \State Set \(S\leftarrow S_o\cup S_h\cup S_p\).
    \State Extract authors and key points of actions and results across the execution according to the entry inclusion rules in Section~\ref{sec:appendix-records}. Arrange them chronologically to form the trace overview.
    \State Construct a separate detailed record for each step in \(S\), ordered chronologically and equipped with content references. Add step references for output steps and production steps.
    \State Assemble the judgment prompt as described in Section~\ref{sec:appendix-judgment} and use a single model call to identify the decisive error step \(\hat{s}\) from \(S\).
    \State \Return \(\hat{s}\) and its author.
  \end{algorithmic}
\end{algorithm}

\subsection{Organization of Execution Records}
\label{sec:appendix-records}

Content can be reused across steps, but each decision involving it has its own execution context. Attribution requires comparing earlier and later processing of shared content while retaining evidence about each execution's task, information, and actions. We organize detailed records of the formation steps accordingly and provide references across steps. The trace overview adds information about execution progress outside the detailed records, allowing local processing to be checked against subsequent use, rejection, or correction.

\paragraph{Detailed records and level of detail.}
The same content may be produced by one agent and received and processed by another, while these executions may have different requirements and access to different information. We therefore retain a separate record for each formation step, ordered chronologically and labeled with its original index and author, presenting its reasoning, action, and tool outputs or environment feedback. These records keep each execution as a distinct object of assessment. They allow the model to use relevant requirements and inputs to check whether a particular action was erroneous and whether the error was introduced earlier or arose in later interpretation or processing.

Actions directly show the processing performed at a step and are not truncated for length. Overly long reasoning, tool outputs, and environment feedback are truncated to the lengths in Table~\ref{tab:params}, with omissions marked. We retain the beginning of reasoning on LongRCA and both the beginning and end of reasoning on Who\&When. For overly long tool outputs or environment feedback, we retain both ends on both benchmarks. On Who\&When, the length limit is applied separately to each returned observation. These choices control the presentation of lengthy text while retaining actions for inspection. Repeated content is organized using the content references described below.

\paragraph{Preprocessing tool outputs and environment feedback.}
Tool outputs and environment feedback may contain task content alongside execution environment messages and formatting fields. In Who\&When records, we remove blank lines, execution environment warnings, and stack-location lines, and process repeated or formatting-related fields in webpage metadata while retaining page addresses and descriptions. These steps reduce the presentation of environment formatting and repeated fields, making the returned content easier to read and check.

\paragraph{Content references.}
The same content may recur in records of its production, handoff, and subsequent processing, providing a shared point of comparison across executions. We share repeated text through references while retaining a separate record for each execution. This allows the model to examine different processing of the shared content without repeatedly reading the same lengthy text in every record.

Specifically, we index identical word sequences in the task description and previously presented detailed records. We match 8-word sequences and mark word positions in the current record that are covered by earlier text. Consecutive covered positions are merged into contiguous regions. Each region of at least 12 words is replaced with a reference such as \texttt{repeats content shown at step i} or a reference to the task description. This minimum covered length corresponds to the content-reference parameter in Table~\ref{tab:params}. If a region has matches from multiple sources, the reference points to the primary matching source that covers the most word positions in that region. Both benchmarks apply content references to actions and tool outputs or environment feedback. The same rules also apply to the reasoning field on Who\&When.

These references provide pointers to relevant original text, allowing the model to compare how shared content was interpreted and processed in different records while reducing repeated presentation. The model uses the records to compare and assess paraphrases and modifications. It also uses execution context to assess actual content use and the role of each instance of processing.

\paragraph{Step references.}
Assessing whether the current processing is appropriate may require revisiting requirements, inputs, or earlier processing from other agents. We reuse the execution links identified during step expansion. For each step \(i\in S_o\cup S_h\), we obtain

\[
R(i)=\{j\in P(i)\cap S\mid a_j\ne a_i\}.
\]

The original index and author of each step in \(R(i)\) are listed in the detailed record of step \(i\). This provides explicit pointers for checking the current action against relevant requirements, inputs, and earlier processing, helping distinguish erroneous requirements or inputs from errors in the current execution's processing. Content references provide shared text for comparison, and step references identify where to revisit related executions. Together with separate records, they allow related executions to be examined together and each decision to be assessed in its own context.

\paragraph{Trace overview.}
A local error's role in the final failure depends on subsequent processing. Content use, rejection, or correction may occur outside the formation steps. To capture these developments, we extract key points of actions and tool outputs or environment feedback across the complete execution and arrange brief entries in the original chronological order. Each entry retains its original step index and author and presents brief key points from the recorded content. On Who\&When, the overview includes only decision steps with action content. On LongRCA, it includes all decision steps, with brief records also provided for steps without action content.

The key points of an action are extracted according to its recorded format and summarize the processing performed at that step, such as the called tool and its arguments, the beginning of a code block, or the recipient and instructions of a dispatch. The key points of a result provide brief outcome indicators and selected returned fields, such as an exit code, a webpage's title and address, or the opening text of a result and the amounts and quantities it states. When longer action key points recur for the same author, we use pointers to their first corresponding entries. This preserves the recurrence while reducing repeated presentation. It helps the model connect earlier and later attempts with execution progress and check whether execution recovered or an earlier error continued to affect subsequent processing.

Detailed records support examination of specific executions, while the overview provides overall progress as context. The model can use both to examine where errors were introduced, how they changed, and what role they played in the final failure. The answer remains restricted to the formation steps \(S\).

\subsection{Judgment Prompts and Answers}
\label{sec:appendix-judgment}

The judgment prompt combines task and outcome information with the trace overview, detailed records, judgment criteria, and answer format. It guides the model to examine related executions together and assess each action in its own context. Judgment prompts for the two benchmarks are presented in Appendix~\ref{sec:appendix-prompts}.

\paragraph{Input arrangement and outcome information.} The task description specifies the original task's objective. On Who\&When, the reference answer provides the expected task outcome. On LongRCA, we extract the system's own statement about its final outcome from the trace and check it against the execution records. Outcome information provides a reference for retrospective analysis, while each action is assessed against the agent's task requirements and the information available at that step. The input protocols are described in Appendix~\ref{sec:appendix-protocols}.

Following the task and outcome information, the overview presents overall execution progress and the detailed records provide evidence about specific actions and their inputs. Content references and step references connect relevant records for comparison. This arrangement allows the model to examine local processing alongside earlier requirements and inputs, as well as subsequent use or correction.

\paragraph{Judgment criteria and output.} The prompts direct the model to assess whether a step's own action was erroneous and examine its role in the final failure through related earlier and later processing. The target is the earliest mistake that was decisive for the final failure. Judgment distinguishes the introduction of an error from later propagation or reporting, as well as from errors that were corrected and exploratory actions from which the execution recovered. Judgment also considers the content actually used in subsequent execution and its outcomes to assess whether a correction changed the error's role in the final failure. By comparing earlier processing, later revisions, and the final outcome, the model assesses the error's role across the execution. On Who\&When, the model also considers how correcting the action would have affected the final outcome, using the reference answer and subsequent execution as evidence.

The model makes a joint attribution judgment in a single call, selecting a decisive error step from the formation steps \(S\) and returning its original index followed by a brief explanation. We use the selected step's author as the predicted responsible agent.

\section{Benchmarks}
\label{sec:appendix-benchmarks}

The two benchmarks cover different collaboration configurations, task domains, and trace lengths, providing a basis for evaluating CoLFA's localization performance and token cost under different execution conditions. This section supplements the benchmark sources and data statistics in Section~\ref{sec:exp-setup} and explains the attribution annotations and task information provided to each method.

\subsection{Evaluation Coverage and Data Statistics}
\label{sec:appendix-coverage-statistics}

\paragraph{Evaluation coverage and sources.}
Collaboration configurations in multi-agent systems affect how requirements and intermediate results are transferred and where related processing occurs. Trace length affects the amount of recorded material that needs to be linked and checked. Who\&When collects failed executions on GAIA and AssistantBench tasks. Its Algorithm-Generated (Alg.) split contains 126 traces produced by teams constructed by CaptainAgent, and its Hand-Crafted (Hand.) split contains 58 traces produced by Magentic-One~\citep{zhang2025agentcausestaskfailures}. The two splits respectively cover broadcast group chats and collaboration in which an orchestrator assigns work to specialized agents, capturing different ways of connecting executions.

LongRCA contains 1,140 failed traces from five sources, covering code repair, terminal operations, travel planning, service-tool use, and web tasks. Its systems include fixed-role teams, group-chat collaboration, and sequential workflows. The benchmark also includes single-agent executions~\citep{longrca}. LongRCA has a greater mean trace length than either split of Who\&When and covers multiple system organizations, providing a broader range for comparing localization performance and input cost under different execution conditions. Expansion scopes for the collaboration configurations are described in Appendix~\ref{sec:appendix-scope}. Settings for the formation-step coverage and input-size analyses are provided in Appendix~\ref{sec:appendix-analysis}.

We use all failed execution traces released by both benchmarks and their corresponding attribution annotations for evaluation. Table~\ref{tab:benchmarks} reports the dataset sizes and original trace lengths.

\begin{table}[!htbp]
  \centering
  \footnotesize
  \setlength{\tabcolsep}{0pt}
  \renewcommand{\arraystretch}{1.0}
  \setlength{\aboverulesep}{0.3ex}
  \setlength{\belowrulesep}{0.3ex}
  \setlength{\cmidrulesep}{0.2ex}
  \begin{tabular*}{\columnwidth}{@{\extracolsep{\fill}}l N{1{,}140} N{000.0} N{000.0}@{}}
    \toprule
    & \hc{Traces} & \hc{Steps} & \hc{Tokens (K)} \\
    \midrule
    \multicolumn{4}{@{}l}{\textit{Who\&When}} \\
    Alg. & 126 & 8.7 & 3.0 \\
    Hand. & 58 & 51.6 & 16.9 \\
    \midrule
    \multicolumn{4}{@{}l}{\textit{LongRCA}} \\
    TravelPlanner & 685 & 165.9 & 48.4 \\
    WebArena Verified & 177 & 58.8 & 31.0 \\
    SWE-bench Pro & 128 & 285.0 & 118.7 \\
    VitaBench & 108 & 54.6 & 10.0 \\
    Terminal-Bench 2 & 42 & 272.1 & 95.8 \\
    \midrule
    All sources & 1{,}140 & 156.0 & 51.7 \\
    \bottomrule
  \end{tabular*}
  \caption{Benchmark statistics. Step and token counts are means per original execution trace.}
  \label{tab:benchmarks}
\end{table}

\paragraph{Statistical conventions.}
A step in Table~\ref{tab:benchmarks} is one record in the original execution trace, including tool and environment records. Decision steps retain the original indices to preserve the correspondence between predicted positions and attribution annotations. We count trace tokens with the DeepSeek-V4-Flash tokenizer using the content of each original record to describe the text size of the original traces. Each method's actual token consumption is measured from API usage. The accounting conventions and breakdowns are provided in Appendix~\ref{sec:token-breakdown}.

\subsection{Annotations and Input Protocols}
\label{sec:appendix-protocols}

\paragraph{Attribution annotations.}
Who\&When annotates the responsible agent and the earliest erroneous step whose correction would turn task failure into success~\citep{zhang2025agentcausestaskfailures}. LongRCA annotates the earliest step that introduces an error leading to the final failure, provided the error is not subsequently corrected. Earlier errors that were corrected and later steps that merely act on, propagate, or reveal an existing error are not treated as the root cause. For handoffs, the instruction step is annotated when the instruction already contains the decisive error. The recipient's step is annotated when the recipient erroneously deviates from the instruction or introduces a new decisive error~\citep{longrca}. These criteria define the error to be localized and its relationship to subsequent processing.

LongRCA annotates the responsible agent and root-cause step separately. CoLFA predicts the responsible agent as the author of the selected step. Both predictions are scored against their respective annotations. Scoring details are provided in Appendix~\ref{sec:appendix-results}.

\paragraph{Input protocols.}
Each method receives the task description and execution trace for the evaluations in which it participates. On Who\&When, each method also receives the expected answer to the original task, namely the task reference answer, following the benchmark's With Ground Truth setting~\citep{zhang2025agentcausestaskfailures}. LongRCA does not release reference answers, and none of the methods evaluated on it receives them. The responsible agent and error step for the current case are attribution annotations used only for scoring. These protocols specify the task information available to each method on each benchmark. Appendix~\ref{sec:appendix-judgment} describes how outcome information is organized in CoLFA's judgment prompt.

\section{Baselines}
\label{sec:appendix-baselines}

\paragraph{Implementation and key adaptations.}
Within each evaluation setting, we rerun the baselines using the same backbone and the task information specified in Appendix~\ref{sec:appendix-benchmarks}. For methods with public implementations, we build on those implementations. For ECHO and RAFFLES, complete runnable implementations are unavailable, so we reproduce them using the algorithms, prompts, and implementation descriptions in their papers.

\paragraph{Evaluation scope.}
DCFA's search requires a task reference answer, so we evaluate it only on Who\&When, which provides this information. For CHIEF, the retrieval resources used for task decomposition in this reproduction come from GAIA and AssistantBench, matching Who\&When's task sources. We therefore also restrict its evaluation to Who\&When.

\paragraph{Base generation settings.}
For each backbone, participating methods use common base generation settings, with ECHO retaining sampling at the temperature required by its method. DeepSeek-V4.1-Flash runs with thinking disabled and temperature 0. GLM-5.3-Flash uses low reasoning effort and greedy decoding. Claude Sonnet 5.5 runs without extended thinking and uses default sampling.

\section{Complete Results}
\label{sec:appendix-results}

\label{sec:appendix-scoring}

\paragraph{Scoring.}
The responsible agent and decisive error step are scored separately. Agent names are compared after normalization, and the predicted step index must exactly match the annotated index. Accuracy is calculated over all evaluated execution traces.

\subsection{Token Breakdown}
\label{sec:token-breakdown}

\begin{table*}[!t]
  \centering
  \footnotesize
  \setlength{\tabcolsep}{0pt}
  \renewcommand{\arraystretch}{1.0}
  \setlength{\aboverulesep}{0.3ex}
  \setlength{\belowrulesep}{0.3ex}
  \setlength{\cmidrulesep}{0.2ex}
  \begin{tabular*}{\textwidth}{@{\extracolsep{\fill}}l@{\hspace{4pt}}N{000{,}000}N{00{,}000}N{000{,}000}@{\hspace{10pt}}N{000{,}000}N{00{,}000}N{000{,}000}@{}}
    \toprule
    & \multicolumn{3}{c}{\textbf{DeepSeek-V4.1-Flash}} & \multicolumn{3}{c}{\textbf{GLM-5.3-Flash}} \\
    \cmidrule(lr){2-4} \cmidrule(lr){5-7}
    Method & \hc{Input} & \hc{Output} & \hc{Total} & \hc{Input} & \hc{Output} & \hc{Total} \\
    \midrule
    \multicolumn{7}{@{}l}{\textit{Who\&When}} \\
    All-at-Once & 7{,}881 & 490 & 8{,}372 & 7{,}829 & 350 & 8{,}179 \\
    A2P & 8{,}125 & 182 & 8{,}308 & 8{,}075 & 360 & 8{,}435 \\
    CORRECT & 21{,}025 & 1{,}707 & 22{,}732 & 19{,}480 & 1{,}133 & 20{,}614 \\
    Step-by-Step & 109{,}979 & 1{,}092 & 111{,}071 & 31{,}188 & 767 & 31{,}955 \\
    Binary Search & 14{,}684 & \textbf{7} & 14{,}691 & 15{,}104 & \textbf{175} & 15{,}279 \\
    ECHO & 52{,}465 & 8{,}419 & 60{,}884 & 52{,}091 & 6{,}391 & 58{,}482 \\
    RAFFLES & 33{,}359 & 1{,}271 & 34{,}630 & 33{,}034 & 1{,}127 & 34{,}160 \\
    DeepDebug & 13{,}181 & 564 & 13{,}745 & 13{,}298 & 917 & 14{,}215 \\
    DCFA & 92{,}341 & 14{,}721 & 107{,}062 & 128{,}548 & 12{,}627 & 141{,}175 \\
    CHIEF & 66{,}741 & 6{,}158 & 72{,}900 & 63{,}224 & 6{,}077 & 69{,}302 \\
    \specialrule{\lightrulewidth}{\aboverulesep}{0pt}
    \shaderow{colfarow}
    \shadestrut \textbf{\textcolor{colfatext}{\method}} & \textbf{6{,}724} & 211 & \textbf{6{,}936} & \textbf{6{,}643} & 417 & \textbf{7{,}060} \\
    \specialrule{\lightrulewidth}{0pt}{\belowrulesep}
    \multicolumn{7}{@{}l}{\textit{LongRCA}} \\
    All-at-Once & 52{,}637 & 183 & 52{,}820 & 51{,}516 & 788 & 52{,}303 \\
    A2P & 53{,}445 & 192 & 53{,}637 & 52{,}337 & 593 & 52{,}930 \\
    CORRECT & 112{,}862 & 1{,}608 & 114{,}471 & 108{,}594 & 1{,}703 & 110{,}296 \\
    Step-by-Step & 943{,}739 & 6{,}296 & 950{,}036 & 364{,}049 & 3{,}788 & 367{,}837 \\
    Binary Search & 130{,}340 & \textbf{13} & 130{,}353 & 114{,}113 & \textbf{555} & 114{,}668 \\
    ECHO & 336{,}848 & 10{,}519 & 347{,}367 & 329{,}722 & 7{,}859 & 337{,}581 \\
    RAFFLES & 214{,}328 & 1{,}556 & 215{,}885 & 210{,}505 & 1{,}803 & 212{,}308 \\
    DeepDebug & 70{,}637 & 673 & 71{,}310 & 70{,}369 & 1{,}559 & 71{,}928 \\
    \specialrule{\lightrulewidth}{\aboverulesep}{0pt}
    \shaderow{colfarow}
    \shadestrut \textbf{\textcolor{colfatext}{\method}} & \textbf{22{,}002} & 401 & \textbf{22{,}403} & \textbf{21{,}312} & 690 & \textbf{22{,}002} \\
    \specialrule{\heavyrulewidth}{0pt}{\belowbottomsep}
  \end{tabular*}
  \caption{Token consumption per execution trace, broken down into input and output. Bold indicates the lowest value in each column for each benchmark. Input and output may not sum to the total because of rounding.}
  \label{tab:tokens}
\end{table*}

Table~\ref{tab:tokens} breaks down the average token consumption reported in Tables~\ref{tab:ww} and~\ref{tab:lr} into input and output to examine the composition of cost across the complete attribution process. Using API-reported usage, we sum input and output tokens across all language model calls within one complete execution of each method and report the mean per trace. For GLM-5.3-Flash, output tokens include reasoning tokens. Token consumption from embedding requests is excluded. Based on the table values, input tokens account for more than 86\% of total consumption for every method, showing that cumulative input constitutes most attribution cost.

CoLFA achieves the lowest total token consumption in all four main evaluation settings. Compared with the baseline with the highest exact step-localization accuracy in each setting, CoLFA reduces total consumption by 66\% to 90\%, with lower consumption of both input and output tokens. The breakdown further shows that reduced input consumption accounts for approximately 90.5\% to 99.4\% of the total savings. Overall savings thus primarily come from the dominant input component and are accompanied by lower output consumption. Together, the breakdown and the localization results in the main text show that CoLFA achieves higher exact step-localization accuracy with lower input and output costs across the complete attribution process.

The cost breakdown in Table~\ref{tab:tokens} and the localization results in the main text jointly show that CoLFA achieves higher exact step-localization accuracy while substantially reducing cumulative input. Cumulative input measures the amount of material processed during attribution, while the support this material provides for judgment also depends on how it is organized. For a given failed execution record, rereading the same history and repeatedly presenting the same original text increase consumption without necessarily adding evidence for distinguishing errors. Attribution in multi-agent systems requires checking relationships among requirements, content production, and subsequent processing, and distinguishing the role of each decision within these relationships. Organizing which steps are examined and how their content is presented can bring scattered records together as evidence that can be jointly examined and compared, while controlling detailed expansion and repeated input. This organization can therefore reduce text processing and improve support for judgment in the same process, providing a common basis for lower overall cost and higher localization accuracy. The contributions of the two components are further examined through the controlled ablations in Appendix~\ref{sec:appendix-ablation}.

\subsection{Complete Ablation}
\label{sec:appendix-ablation}

\begin{table*}[!t]
  \centering
  \footnotesize
  \setlength{\tabcolsep}{0pt}
  \renewcommand{\arraystretch}{1.0}
  \setlength{\aboverulesep}{0.3ex}
  \setlength{\belowrulesep}{0.3ex}
  \setlength{\cmidrulesep}{0.2ex}
  \begin{tabular*}{\textwidth}{@{\extracolsep{\fill}}l@{\hspace{4pt}}rr@{\hspace{3pt}}rr@{\hspace{3pt}}rr@{\hspace{3pt}}r@{\hspace{10pt}}rrrrr@{}}
    \toprule
    & \multicolumn{7}{c}{\textbf{Who\&When}} & \multicolumn{5}{c}{\textbf{LongRCA}} \\
    \cmidrule(lr){2-8} \cmidrule(lr){9-13}
    & \multicolumn{2}{c}{Alg.} & \multicolumn{2}{c}{Hand.} & \multicolumn{2}{c}{Overall} & & & & & & \\
    \cmidrule(lr){2-3} \cmidrule(lr){4-5} \cmidrule(lr){6-7}
    Variant & \numhead{00.00}{Agent} & \numhead{00.00}{Step} & \numhead{00.00}{Agent} & \numhead{00.00}{Step} & \numhead{00.00}{Agent} & \numhead{00.00}{Step} & \numhead{00.0}{Tok.\,(K)} & \numhead{00.00}{Role} & \numhead{00.00}{Exact} & \numhead{00.00}{$\pm$5} & \numhead{00.00}{MAE} & Tok.\,(K) \\
    \midrule
    \multicolumn{13}{@{}l}{\textit{DeepSeek-V4.1-Flash}} \\
    \shaderow{colfarow}
    \shadestrut \textbf{\textcolor{colfatext}{\method}} & \textbf{65.87} & \textbf{52.38} & \textbf{62.07} & \textbf{27.59} & \textbf{64.67} & \textbf{44.57} & \textbf{6.9} & \textbf{51.58} & \textbf{21.84} & \textbf{32.81} & \textbf{35.54} & \textbf{22.4} \\
    w/o Expansion & 61.11 & 46.83 & 56.90 & 22.41 & 59.78 & 39.13 & 7.6 & 47.02 & 19.65 & 30.53 & 37.91 & 32.9 \\
    w/o Organization & 61.90 & 47.62 & 58.62 & 24.14 & 60.87 & 40.22 & 7.5 & 49.47 & 19.30 & 29.47 & 37.80 & 30.1 \\
    w/o Both & 57.94 & 44.44 & 55.17 & 20.69 & 57.07 & 36.96 & 8.4 & 45.26 & 17.89 & 27.02 & 39.45 & 52.0 \\
    \midrule
    \multicolumn{13}{@{}l}{\textit{GLM-5.3-Flash}} \\
    \shaderow{colfarow}
    \shadestrut \textbf{\textcolor{colfatext}{\method}} & \textbf{68.25} & \textbf{54.76} & \textbf{62.07} & \textbf{29.31} & \textbf{66.30} & \textbf{46.74} & \textbf{7.1} & \textbf{52.28} & \textbf{23.60} & \textbf{36.23} & \textbf{35.42} & \textbf{22.0} \\
    w/o Expansion & 64.29 & 50.79 & 56.90 & 24.14 & 61.96 & 42.39 & 7.7 & 49.82 & 20.70 & 34.04 & 36.70 & 32.4 \\
    w/o Organization & 65.08 & 51.59 & 56.90 & 25.86 & 62.50 & 43.48 & 7.7 & 48.07 & 20.35 & 33.33 & 36.77 & 29.9 \\
    w/o Both & 60.32 & 48.41 & 51.72 & 18.97 & 57.61 & 39.13 & 8.5 & 46.32 & 19.30 & 29.12 & 40.14 & 51.0 \\
    \bottomrule
  \end{tabular*}
  \caption{Complete ablation results for CoLFA by split, with strict matching on Who\&When. Accuracies are percentages, and MAE is measured in steps. Bold indicates the best value for each backbone.}
  \label{tab:ablation-full}
\end{table*}

The first seven metric columns report Who\&When (WW), and the last five report LongRCA (LR).

Table~\ref{tab:ablation-full} uses the four main evaluation settings from the main text and compares the four combinations obtained by retaining or removing each of the two components. It provides results for the Who\&When splits and the complete set of metrics for both benchmarks to examine how each component and their combination contribute to localization performance and token cost. Within each setting, all variants use the same task information, base judgment prompt, backbone, and generation settings. Only input organization and the corresponding variant descriptions are changed, and each execution trace is read only once.

Table~\ref{tab:ablation-full} shows that both components improve localization on both Who\&When splits. Compared with w/o Expansion and w/o Organization under both backbones, CoLFA improves Step by 3.17\% to 5.56\% and Agent by 3.17\% to 4.76\% on Alg. On Hand., both Step and Agent improve by 3.45\% to 5.17\%. At the same time, total token consumption for Who\&When as a whole decreases by approximately 8\% to 9\%. Compared with w/o Both, Agent and Step gains are larger in both splits, and total consumption decreases by approximately 17\%. Each component improves localization and reduces consumption even when the other is already retained. This shows their complementary value in supporting judgment and controlling cost, as well as the greater gains from their joint use.

Table~\ref{tab:ablation-full} shows that on LongRCA, CoLFA achieves both higher exact step-localization accuracy and lower mean localization error. Compared with w/o Expansion and w/o Organization under both backbones, Exact increases by 2.19\% to 3.25\%, while \(\pm\)5 increases and MAE decreases. These changes extend the gains in exact step localization reported in the main text to accuracy within a five-step tolerance and mean error. They indicate that both levels of organization help the model distinguish specific error locations among related executions, providing consistent evidence for the ablation findings across different error scales. These localization improvements are accompanied by lower overall cost. Total token consumption is approximately 26\% to 32\% lower than with the two ablation variants that remove one component and approximately 57\% lower than with w/o Both.

\subsection{Paired Bootstrap Analysis of Main Results}
\label{sec:appendix-bootstrap}

To further assess the reliability of the exact step-localization accuracy differences in the main experiments, we perform trace-level paired nonparametric bootstrap analysis across the four settings. We use all 184 traces from Who\&When and all 1,140 traces from LongRCA, comparing CoLFA with the most accurate baseline in each setting, determined from the main results before bootstrap analysis. In each replicate, we sample complete execution traces with replacement from the test set, drawing as many traces as in the original sample, and use the same sampled indices to calculate the accuracy difference between the two methods. We perform 10,000 replicates and use the percentile method to construct 95\% confidence intervals from the 2.5th and 97.5th percentiles of the resampled differences.

\begin{table*}[!t]
  \centering
  \footnotesize
  \setlength{\tabcolsep}{0pt}
  \renewcommand{\arraystretch}{1.0}
  \setlength{\aboverulesep}{0.3ex}
  \setlength{\belowrulesep}{0.3ex}
  \setlength{\cmidrulesep}{0.2ex}
  \begin{tabular*}{\textwidth}{@{\extracolsep{\fill}}l@{\hspace{4pt}}lrl@{\hspace{3pt}}rrr@{}}
    \toprule
    Dataset & Backbone & \numhead{\textbf{00.00}}{\textcolor{colfatext}{CoLFA (\%)}} & \multicolumn{2}{c}{Best Baseline (\%)} & \numhead{$+0.00$}{$\Delta$ (\%)} & \numhead{$[-0.00, 0.00]$}{95\% CI (\%)} \\
    \midrule
    Who\&When & DeepSeek-V4.1-Flash & \textbf{44.57} & CORRECT & (41.30) & $+3.26$ & $[-1.72, 9.78]$ \\
    Who\&When & GLM-5.3-Flash & \textbf{46.74} & CORRECT & (44.57) & $+2.17$ & $[-2.80, 8.15]$ \\
    \midrule
    LongRCA & DeepSeek-V4.1-Flash & \textbf{21.84} & CORRECT & (18.86) & $+2.98$ & $[0.44, 5.53]$ \\
    LongRCA & GLM-5.3-Flash & \textbf{23.60} & RAFFLES & (21.75) & $+1.84$ & $[0.28, 3.51]$ \\
    \bottomrule
  \end{tabular*}
  \caption{Paired bootstrap comparison of CoLFA with the most accurate baseline in each setting. $\Delta$ is CoLFA accuracy minus baseline accuracy before rounding. Differences and confidence intervals are reported in \%. Bold marks the higher accuracy in each setting.}
  \label{tab:paired-bootstrap}
\end{table*}

CoLFA has a positive exact step-localization accuracy difference relative to the most accurate baseline in all four main settings. On LongRCA, which has longer execution traces on average, the 95\% confidence intervals lie above zero under both backbones, providing further statistical support for CoLFA's exact step-localization advantage on this benchmark. On Who\&When, the accuracy gains under the two backbones are 3.26\% and 2.17\%, similar in magnitude to those on LongRCA, but the confidence intervals are wider and include zero. Who\&When has a smaller test set, making estimates of accuracy differences more sensitive to variation from trace resampling, consistent with the observed wider confidence intervals.

\section{Analysis Details}
\label{sec:appendix-analysis}

This section examines the execution records and organized attribution inputs to further assess the design rationale for using collaboration cues to support attribution, as discussed in Section~\ref{sec:exp-analysis}. The analyses consider coverage of decisive error steps by formation steps, the trade-off between coverage and size as the specificity threshold changes, the textual basis for shared points of comparison, how attribution input size varies with execution trace length, and the localization effect of step selection. The statistics are based on the 1,032 multi-agent execution traces from LongRCA. The coverage analysis additionally includes the 58 execution traces from Who\&When Hand-Crafted.

\subsection{Coverage of Formation Steps}
\label{sec:appendix-formation-coverage}

Decisive errors can be introduced at steps that differ from the final output in both author and execution position. Among cases where the annotated error is a decision step, 65.3\% of decisive error steps in LongRCA's multi-agent traces and 36.8\% in Who\&When Hand. are authored by agents other than the output agent (Table~\ref{tab:coverage-author}). Figure~\ref{fig:error-position} further shows that error positions vary by source, with errors tending to occur earlier in SWE-bench Pro and Terminal-Bench 2. These distributions provide empirical support for linking the final output with related earlier executions so that they can be examined together.

Table~\ref{tab:coverage-author} shows that, with the same number of selected steps, formation steps achieve higher coverage of error steps authored by agents other than the output agent. All decision steps of the output agent are retained by construction. Analyzing errors by other agents separately therefore checks whether the records added by tracing across agents contain positions that need to be examined. When the number of selected decision steps is matched for each trace, coverage on LongRCA is 72.4\%, compared with 54.8\% for the most recent steps and an expected coverage of 49.6\% for random selection. The gap is wider on SWE-bench Pro and Terminal-Bench 2, where errors occur earlier, with absolute coverage gains of 52.3\% and 28.6\% over the most recent steps, respectively. Errors introduced in earlier executions may lead to final failure only after subsequent processing, so temporal proximity to the output is an incomplete indicator of a record's value for attribution checks. These results show that organizing the scope through execution links can retain more annotated error positions across agents and execution stages, supporting examination of earlier processing together with the final result.

\begin{table}[!htbp]
  \centering
  \footnotesize
  \setlength{\tabcolsep}{0pt}
  \renewcommand{\arraystretch}{1.0}
  \setlength{\aboverulesep}{0.3ex}
  \setlength{\belowrulesep}{0.3ex}
  \setlength{\cmidrulesep}{0.2ex}
  \begin{tabular*}{\columnwidth}{@{\extracolsep{\fill}}l N{00.0} N{00.0} N{00.0} N{00.0}@{}}
    \toprule
    & & \multicolumn{3}{c@{}}{Coverage} \\
    \cmidrule(l){3-5}
    Source & \hc{Other} & \hc{\textcolor{colfatext}{Formation}} & \hc{Recent} & \hc{Random} \\
    \midrule
    SWE-bench Pro & 35.2 & 68.2 & 15.9 & 35.5 \\
    Terminal-Bench 2 & 52.5 & 52.4 & 23.8 & 38.0 \\
    TravelPlanner & 67.3 & 71.7 & 55.9 & 49.4 \\
    WebArena Verified & 81.9 & 78.6 & 67.6 & 56.2 \\
    \midrule
    \shortstack[l]{LongRCA (four\\sources combined)} & 65.3 & 72.4 & 54.8 & 49.6 \\
    Who\&When Hand. & 36.8 & 95.2 & 81.0 & 89.9 \\
    \bottomrule
  \end{tabular*}
  \caption{Coverage of decisive error steps authored by agents other than the output agent. Values are percentages. Other is the share of these errors among cases whose annotated decisive error step is a decision step.}
  \label{tab:coverage-author}
\end{table}

\subsection{Threshold Sensitivity}
\label{sec:appendix-threshold}

Table~\ref{tab:theta} shows that the proportion of decision steps selected as formation steps and error coverage both decrease as content selection becomes stricter. \(\theta\) is the minimum content specificity defined in Appendix~\ref{sec:appendix-link-rules}. We keep the handoff rules and expansion procedure unchanged and vary only this threshold. As \(\theta\) increases from 1.0 to 1.8, the mean proportion of all decision steps retained as formation steps decreases from 50.0\% to 47.7\%, and error coverage decreases from 79.7\% to 77.2\%. Coverage here is calculated over all 1,032 multi-agent execution traces from LongRCA. These changes reflect the trade-off between retaining more content correspondence cues and controlling the scope of examination.

After matching the number of selected decision steps for each trace at each threshold, the absolute coverage advantage of formation steps over the most recent steps remains between 26.1\% and 27.1\%. The comparison sets vary with the number of formation steps, showing that the advantage is observed under both wider and narrower examination scopes. The threshold thus adjusts the trade-off between coverage and size, while the coverage advantage of using execution links to select the scope remains similar within the tested range.

\begin{table}[!htbp]
  \centering
  \footnotesize
  \setlength{\tabcolsep}{0pt}
  \renewcommand{\arraystretch}{1.0}
  \setlength{\aboverulesep}{0.3ex}
  \setlength{\belowrulesep}{0.3ex}
  \setlength{\cmidrulesep}{0.2ex}
  \begin{tabular*}{\columnwidth}{@{\extracolsep{\fill}}l N{00} N{00.0} N{00.0} N{00.0} N{00.0}@{}}
    \toprule
    $\theta$ & \hc{Share limit} & \hc{$|S|/|V|$} & \hc{Coverage} & \hc{Recent} & \hc{Gap} \\
    \midrule
    1.0 & 37 & 50.0 & 79.7 & 52.6 & 27.1 \\
    1.2 & 30 & 49.7 & 79.4 & 52.4 & 27.0 \\
    \shaderow{colfarow}
    \shadestrut \textbf{\textcolor{colfatext}{1.4}} & \textbf{25} & \textbf{49.3} & \textbf{78.9} & \textbf{52.1} & \textbf{26.7} \\
    1.6 & 20 & 48.7 & 78.3 & 51.6 & 26.7 \\
    1.8 & 17 & 47.7 & 77.2 & 51.1 & 26.1 \\
    \bottomrule
  \end{tabular*}
  \caption{Sensitivity to the specificity threshold \(\theta\) on multi-agent execution traces from LongRCA. All columns except \(\theta\) are reported in \%. Gap is the absolute coverage difference between formation steps and the same number of most recent decision steps. Bold marks the setting used by CoLFA. Share limit, approximately \(e^{-\theta}\), is the maximum proportion of original records in which a retained item may occur.}
  \label{tab:theta}
\end{table}

\subsection{Textual Basis for Shared Points of Comparison}
\label{sec:appendix-shared-text}

Table~\ref{tab:content-refs} shows that shared text remains widespread within the detailed records of formation steps. Content references appear in 99.9\% of multi-agent traces, and the replaced spans account for 24.0\% of the original character count of actions and tool outputs or environment feedback before truncation. The scope of this statistic shows that shareable repeated text persists within the selected records after the range of steps examined in detail has been narrowed. Different executions in multi-agent collaboration may need to be examined together because of their processing while also containing the same requirements, information, or handed-off content. Retaining these executions and reducing repeated presentation of original text can therefore be achieved together. These findings provide empirical support for organizing content within the selected step range and further explain why organization at both levels can jointly control the amount of attribution material.

In the aggregate statistics, replaced text accounts for 24.7\% of the original characters in actions and 23.4\% in tool outputs or environment feedback, indicating comparable proportions of shareable text in both types of material. For failure attribution, action records support checking what processing each execution performed, while tool outputs or environment feedback provide information for checking that processing and subsequent decisions. Shared spans in actions can support comparisons of specific processing and its execution contexts across records. Shared spans in tool outputs or environment feedback can support checking how existing information was understood and used later. These distributions break down the overall repetition ratio reported in the main text into two types of material used for judgment. They show that shareable original text exists in both the processing and feedback to be examined, providing a common textual basis for reducing repeated presentation while retaining evidence for comparing earlier and later processing.

\begin{table}[!htb]
  \centering
  \footnotesize
  \setlength{\tabcolsep}{0pt}
  \renewcommand{\arraystretch}{1.0}
  \setlength{\aboverulesep}{0.3ex}
  \setlength{\belowrulesep}{0.3ex}
  \setlength{\cmidrulesep}{0.2ex}
  \begin{tabular*}{\columnwidth}{@{\extracolsep{\fill}}l N{00.0} N{00.0} N{00.0} N{00.0} N{00.0}@{}}
    \toprule
    & \multicolumn{5}{c@{}}{Share (\%)} \\
    \cmidrule(l){2-6}
    Source & \hc{Traces} & \hc{Steps} & \hc{Actions} & \hc{\shortstack{Tool outputs\\or\\environment\\feedback}} & \hc{Total} \\
    \midrule
    TravelPlanner & 100 & 18.6 & 7.8 & 28.1 & 20.9 \\
    WebArena Verified & 100 & 56.1 & 62.0 & 61.9 & 62.0 \\
    SWE-bench Pro & 100 & 35.1 & 9.2 & 11.9 & 11.3 \\
    Terminal-Bench 2 & 97.6 & 28.3 & 12.2 & 13.1 & 12.8 \\
    \midrule
    All & 99.9 & 25.2 & 24.7 & 23.4 & 24.0 \\
    \bottomrule
  \end{tabular*}
  \caption{Content references in CoLFA's detailed records on multi-agent execution traces from LongRCA. Traces and Steps report the proportions of traces and formation steps containing content references. The remaining columns report the proportions of original characters replaced by content references in actions, tool outputs or environment feedback, and their total.}
  \label{tab:content-refs}
\end{table}

\subsection{Attribution Input Size}
\label{sec:appendix-input-size}

Table~\ref{tab:input-scale} shows that CoLFA reduces input size more on longer execution traces. We divide traces into five equal-sized groups by the token count of the complete execution trace. Based on the token totals within each group, the attribution inputs are approximately 42\% and 69\% smaller than the complete traces in the shortest and longest groups, respectively. These inputs include instructions, detailed records, and a trace overview covering the entire execution, so the reductions concern the complete judgment prompt. These differences indicate greater room for savings in longer collaboration records and a more pronounced input size advantage for failure attribution on long executions.

This relationship with length is also observed within each source. Linear fits of attribution input token counts against complete trace token counts in log-log coordinates yield slopes from 0.57 to 0.68 across sources, all below 1. Meanwhile, from the shortest to the longest group, the mean proportion of all decision steps included as formation steps decreases from 60.5\% to 39.0\%. The overview covers all decision steps in every group, while detailed records cover a smaller proportion of decisions in longer groups. This indicates a more focused range of detailed presentation relative to the complete execution. The overall correspondence between the share of steps expanded in detail and relative input size provides structural support for CoLFA's approach to setting the level of detail according to what needs to be checked and controlling input size in long collaborative executions.

\begin{table}[!htb]
  \centering
  \footnotesize
  \setlength{\tabcolsep}{0pt}
  \renewcommand{\arraystretch}{1.0}
  \setlength{\aboverulesep}{0.3ex}
  \setlength{\belowrulesep}{0.3ex}
  \setlength{\cmidrulesep}{0.2ex}
  \begin{tabular*}{\columnwidth}{@{\extracolsep{\fill}}l N{72.1--394.2} N{00.0} N{00} N{00}@{}}
    \toprule
    & \multicolumn{2}{c}{Tokens (K)} & \multicolumn{2}{c@{}}{Ratio (\%)} \\
    \cmidrule(lr){2-3} \cmidrule(l){4-5}
    Group & \hc{Trace} & \hc{Input} & \hc{Aggregate} & \hc{Median} \\
    \midrule
    Shortest quintile & 5.8--30.8 & 12.1 & 58 & 61 \\
    Second quintile & 30.8--38.2 & 12.5 & 44 & 36 \\
    Middle quintile & 38.2--51.3 & 22.1 & 50 & 49 \\
    Fourth quintile & 51.3--72.0 & 25.9 & 43 & 44 \\
    Longest quintile & 72.1--394.2 & 34.2 & 31 & 32 \\
    \midrule
    All & 5.8--394.2 & 21.5 & 40 & 44 \\
    \bottomrule
  \end{tabular*}
  \caption{Size of CoLFA's attribution input on multi-agent execution traces from LongRCA, grouped into quintiles by complete execution trace token count. Input reports medians. Aggregate ratio is the ratio of token totals, and Median ratio is the median of per-trace token ratios, both relative to complete execution traces.}
  \label{tab:input-scale}
\end{table}

\subsection{Step Selection and Localization}
\label{sec:appendix-step-selection}

Table~\ref{tab:step-selection} shows that, when the output agent's steps are retained and the number of selected steps is matched for each trace, the steps CoLFA selects through collaboration cues cover more decisive errors and achieve the highest exact step-localization accuracy and the lowest CoP under both backbones. We compare CoLFA with Recent and Random on the 1,032 multi-agent execution traces from LongRCA. All strategies retain the output agent's decision steps and add the same number of steps from other agents. CoLFA adds its production and supply steps, Recent adds the latest decision steps of other agents, and Random samples them uniformly. Unlike the coverage comparison in Figure~\ref{fig:error-position}, the controls here also keep the output agent's steps. Task information, the trace overview, content organization, the judgment prompt, and backbone settings are held constant, so the comparison tests which steps are selected rather than how many.

\begin{table}[!htbp]
  \centering
  \footnotesize
  \setlength{\tabcolsep}{0pt}
  \renewcommand{\arraystretch}{1.0}
  \setlength{\aboverulesep}{0.3ex}
  \setlength{\belowrulesep}{0.3ex}
  \setlength{\cmidrulesep}{0.2ex}
  \begin{tabular*}{\columnwidth}{@{\extracolsep{\fill}}l N{00.0} N{00.0} N{00.00} N{00.0} N{000} N{00.00} N{00.0} N{000}@{}}
    \toprule
    & \multicolumn{2}{c}{Coverage} & \multicolumn{3}{c}{DeepSeek} & \multicolumn{3}{c@{}}{GLM} \\
    \cmidrule(lr){2-3}\cmidrule(lr){4-6}\cmidrule(l){7-9}
    Strategy & \hc{All} & \hc{\shortstack{Other\\agents}} & \hc{Exact} & \hc{Tok.} & \hc{CoP} & \hc{Exact} & \hc{Tok.} & \hc{CoP} \\
    \midrule
    \shaderow{colfarow}
    \shadestrut \textbf{\textcolor{colfatext}{\method}} & \textbf{78.9} & \textbf{72.4} & \textbf{21.12} & 23.3 & \textbf{110} & \textbf{22.97} & 22.9 & \textbf{100} \\
    Recent & 61.5 & 44.8 & 17.93 & 21.2 & 118 & 18.12 & 21.0 & 116 \\
    Random & 57.1 & 37.7 & 17.15 & \textbf{20.3} & 119 & 16.18 & \textbf{20.2} & 125 \\
    \bottomrule
  \end{tabular*}
  \caption{Matched-step attribution on 1,032 multi-agent execution traces from LongRCA. Coverage and Exact are reported in \%, and Tok. and CoP in thousands of tokens. Other agents is the coverage of decisive errors introduced by agents other than the output agent. Coverage depends only on the selected steps and is shared by both backbones. Bold marks the best values.}
  \label{tab:step-selection}
\end{table}

CoLFA covers 78.9\% of decisive errors, compared with 61.5\% for Recent and 57.1\% for Random. All strategies cover the errors introduced by the output agent, so the differences lie in errors introduced by other agents, which account for 65.3\% of decisive errors at decision steps (Table~\ref{tab:coverage-author}). CoLFA covers 72.4\% of these errors, compared with 44.8\% for Recent and 37.7\% for Random. In multi-agent tasks, different agents may introduce errors at different stages of execution, and execution order alone does not reveal which earlier steps relate to the final failure. By tracing how handed-off content was produced and adding the requirements and inputs supplied by other agents, CoLFA brings the executions of other agents that introduce such errors into the examined scope.

Coverage is a prerequisite for correct localization, since the model identifies the decisive error among the selected steps. As CoLFA brings more decisive error steps into the candidate set, its localization accuracy is correspondingly higher. CoLFA reaches an Exact of 21.12\% with DeepSeek, exceeding Recent and Random by 3.20\% and 3.97\%, and 22.97\% with GLM, exceeding them by 4.84\% and 6.78\%. The additional correct localizations also lower its CoP by 6.9\% to 20.1\% relative to the two controls. These results show that organizing the examined scope through collaboration cues brings more decisive error steps into the candidate set and improves final localization accuracy.

\makeatletter\setlength{\@fptop}{0pt}\setlength{\@fpsep}{8pt}\setlength{\@fpbot}{0pt plus 1fil}\makeatother

\begin{figure*}[!t]
\section{Judgment Prompts}
\label{sec:appendix-prompts}

Prompts~\ref{lst:prompt-lr} and~\ref{lst:prompt-ww} present CoLFA's judgment prompts for LongRCA and Who\&When, including the inputs, judgment criteria, and outputs. Input construction is described in Appendix~\ref{sec:appendix-input}.

  \begin{prompttemplate}{Judgment Prompt on LongRCA}
    \label{lst:prompt-lr}
    \input{sections/_prompt_lr}
  \end{prompttemplate}
\end{figure*}

\begin{figure*}[t]
  \begin{prompttemplate}{Judgment Prompt on Who\&When}
    \label{lst:prompt-ww}
    \input{sections/_prompt_ww}
  \end{prompttemplate}
\end{figure*}

\end{document}